\documentclass[journal]{IEEEtran}
\usepackage{cite}

\ifCLASSINFOpdf
  \usepackage[pdftex]{graphicx}
  \graphicspath{{figures/}}
\else
\fi
\usepackage{amsmath}
\usepackage{amssymb}
\usepackage{algorithm}
\usepackage{algpseudocode}

\algnewcommand\algorithmicforeach{\textbf{for each}}
\algdef{S}[FOR]{ForEach}[1]{\algorithmicforeach\ #1\ \algorithmicdo}
\algdef{SE}[DOWHILE]{Do}{doWhile}{\algorithmicdo}[1]{\algorithmicwhile\ #1}%

\algrenewcommand\algorithmicindent{1.0em}

\ifCLASSOPTIONcompsoc
  \usepackage[caption=false,font=normalsize,labelfont=sf,textfont=sf]{subfig}
\else
  \usepackage[caption=false,font=footnotesize]{subfig}
\fi
\usepackage{booktabs,makecell}
\usepackage{siunitx}

\usepackage{multirow}
\usepackage{hhline}
\usepackage[table,xcdraw]{xcolor}

\usepackage{url}

\makeatletter
\let\NAT@parse\undefined
\makeatother
\usepackage{hyperref}  

\usepackage{array}
\newcolumntype{L}[1]{>{\raggedright\arraybackslash}p{#1}}
\newcolumntype{C}[1]{>{\centering\arraybackslash}p{#1}}
\newcolumntype{R}[1]{>{\raggedleft\arraybackslash}p{#1}}

\usepackage{rotating,siunitx}

\usepackage{soul}
\soulregister\cite7
\soulregister\citep7
\soulregister\citet7
\soulregister\ref7
\soulregister\pageref7

\begin{document}

\title{Efficient Match Pair Retrieval for Large-scale UAV Images via Graph Indexed Global Descriptor}

\author{San Jiang,
        Yichen Ma,
        Qingquan Li,
        Wanshou Jiang,
        Bingxuan Guo,
        Lelin Li,
        and Lizhe Wang
\thanks{S. Jiang, Y. Ma, and L. Wang are with the School of Computer Science, China University of Geosciences, Wuhan 430074, China; S. Jiang is also with the Guangdong Laboratory of Artificial Intelligence and Digital Economy (Shenzhen), Shenzhen 518060, China, and with the Hubei Key Laboratory of Intelligent Geo-Information Processing, China University of Geosciences, Wuhan 430078, China. E-mail: \textit{jiangsan}@cug.edu.cn, \textit{mayichen}@cug.edu.cn, \textit{lzwang@cug.edu.cn}. \textit{(Corresponding author: Lizhe Wang})}
\thanks{Q. Li is with the College of Civil and Transportation Engineering, Shenzhen University, Shenzhen 518060, China, and also with the Guangdong Laboratory of Artificial Intelligence and Digital Economy (Shenzhen), Shenzhen 518060, China. E-mail: \textit{liqq}@szu.edu.cn.}
\thanks{W. Jiang and B. Guo are with the State Key Laboratory of Information Engineering in Surveying, Mapping, and Remote Sensing, Wuhan University, Wuhan 430072, China. E-mail: \textit{jws}@whu.edu.cn, \textit{mobilemap}@163.com.}
\thanks{L. Li is with the Provincial Key Laboratory of Geo-information Engineering in Surveying, Mapping and Remote Sensing, Hunan University of Science
and Technology, Xiangtan 411201, China. E-mail: \textit{lilelin}@hnust.edu.cn.}}



\maketitle

\begin{abstract}
SfM (Structure from Motion) has been extensively used for UAV (Unmanned Aerial Vehicle) image orientation. Its efficiency is directly influenced by feature matching. Although image retrieval has been extensively used for match pair selection, high computational costs are consumed due to a large number of local features and the large size of the used codebook. Thus, this paper proposes an efficient match pair retrieval method and implements an integrated workflow for parallel SfM reconstruction. First, an individual codebook is trained online by considering the redundancy of UAV images and local features, which avoids the ambiguity of training codebooks from other datasets. Second, local features of each image are aggregated into a single high-dimension global descriptor through the VLAD (Vector of Locally Aggregated Descriptors) aggregation by using the trained codebook, which remarkably reduces the number of features and the burden of nearest neighbor searching in image indexing. Third, the global descriptors are indexed via the HNSW (Hierarchical Navigable Small World) based graph structure for the nearest neighbor searching. Match pairs are then retrieved by using an adaptive threshold selection strategy and utilized to create a view graph for divide-and-conquer based parallel SfM reconstruction. Finally, the performance of the proposed solution has been verified using three large-scale UAV datasets. The test results demonstrate that the proposed solution accelerates match pair retrieval with a speedup ratio ranging from 36 to 108 and improves the efficiency of SfM reconstruction with competitive accuracy in both relative and absolute orientation.
\end{abstract}


%
\IEEEpeerreviewmaketitle

\section{Introduction}
\label{sec:1}
%
%
%
%

\IEEEPARstart{U}{AV} (Unmanned aerial vehicle) images have become one of the primary data sources for surveying and mapping in photogrammetry and remote sensing (RS). Compared with satellite and aerial-based RS platforms, UAVs have the characteristics of high flexibility, high timeliness, and high resolution \cite{jiang2021unmanned}. UAV images have been widely exploited in various applications, e.g., urban 3D modeling \cite{li2023optimized}, transmission line inspection \cite{jiang2019uav,jiang2017uav}, and precision agriculture management \cite{colomina2014unmanned}. With the increasing endurance of UAV platforms and the explosive usage of multi-camera instruments, efficient image orientation for large-scale UAV images has become one of the most critical modules for photogrammetric systems \cite{jiang2020efficient}.

 SfM (Structure from Motion) has become a well-known technology for recovering camera poses and 3D points without the requirement of their good initial values \cite{schonberger2016structure}. SfM has been extensively adopted in 3D reconstruction \cite{nikolakopoulos2017uav,wischounig2015resource} for both ordered and unordered UAV images. In the workflow of SfM, a view graph is a basic structure to guide feature matching and parameter solving, which is defined as an undirected weighted graph with the vertices and edges indicating images and their overlap relationships \cite{chen2020graph,cui2021view}. Retrieving match pairs is pre-required in view graph construction. The purpose of match pair retrieval is to find overlapped image pairs to guide subsequent feature matching, which increases the reliability and efficiency of SfM reconstruction. Thus, retrieving appropriate match pairs efficiently and accurately becomes one of the core issues in SfM for large-scale UAV images.

In the literature, existing methods for retrieving match pairs can be divided into two categories, i.e., prior knowledge-based and visual similarity-based methods. The former depends on prior information, such as the sequential constraint in data acquisitions \cite{aliakbarpour2015fast,schonberger2014structure} or depends on prior data from onboard POS (Positioning and Orientation System) sensors \cite{jiang2018efficient,xu2016skeletal} to calculate image ground footprints. Although these methods are very efficient, their usage is limited to the special configurations of data acquisition or depends on the precision of the prior data from used RS platforms. Without relying on other auxiliary data, visual similarity-based methods merely use images to calculate similarity scores between two images and determine overlapped match pairs by selecting images with the highest similarity scores. The most commonly used solution is CBIR (Content-Based Image Retrieval). The core idea of CBIR is to encode detected local features, e.g., SIFT (Scale Invariant Feature Transform) \cite{lowe2004distinctive}, into high-dimension vectors, and the problem of retrieving match pairs is then cast as calculating the similarity score between two of these high-dimension vectors \cite{hou2023learning}. In the fields of photogrammetry and computer vision, vocabulary tree \cite{nister2006scalable} based image retrieval has become the most classic method that converts local features into high-dimension BoW (Bag-of-Words) vectors \cite{jiang2022leveraging}.

In vocabulary tree-based image retrieval, the similarity calculation uses an inverted index that establishes the relationship between visual words and corresponding local features \cite{galvez2012bags}. However, building the inverted index is time-consuming for high-resolution and large-size UAV images. On the one hand, high-resolution UAV images lead to tens of thousands of local features from an individual image, which causes high computational costs in searching the nearest visual word via ANN (Approximate Nearest Neighbor) searching; on the other hand, large-volume UAV images requires an extremely large codebook to increase the discriminative ability of aggregated BoW vectors, which causes the millions of vector dimensions and further increases computational costs in ANN searching. In addition, the codebook is usually created offline from public datasets due to the high time costs of generating a large codebook. Thus, this study proposes an efficient and accurate solution for match pair retrieval. The core idea is to adopt a global descriptor for image representation and explore graph indexing for efficient ANN searching of high-dimension vectors. Our main contributions are summarized: (1) An individual codebook is trained online using random selection and scale restriction strategies to reduce image and feature redundancies. (2) Local features of each image are aggregated into a high-dimension global descriptor through a VLAD (Vector of Locally Aggregated Descriptors) aggregation that extremely reduces the number of features and the burden of nearest neighbor searching in image indexing. (3) VLAD descriptors are indexed into an HNSW (Hierarchical Navigable Small World) based graph structure for the ANN (Approximate Nearest Neighbor) searching, and match pairs are retrieved using an adaptive threshold selection strategy, which is used to create a view graph to for divide-and-conquer based parallel SfM reconstruction. (4) The performance of the proposed solution is verified by using large-scale UAV images and compared with other well-known software packages.

The structure of this study is organized as follows. Section \ref{sec:2} gives a literature review of match pair retrieval and nearest neighbor searching. Section \ref{sec:3}presents detailed procedures of the proposed match pairs retrieval algorithm and the workflow of the parallel SfM solution. Section \ref{sec:4} conducts a comprehensive evaluation and comparison using UAV datasets. Finally, Section \ref{sec:5} gives the conclusion of this study and improvements for future research.

\section{Related work}
\label{sec:2}


This study focuses on match pair retrieval to improve the efficiency of SfM reconstruction. Thus, this section reviews match pair selection and nearest neighbor searching.

\subsection{Prior knowledge-based methods}
\label{sec:2.1}

 For photogrammetric data acquisition, there are usually two categories of prior knowledge, i.e., the configuration for data acquisition and the auxiliary data from onboard sensors. For the former, image match pairs are usually obtained according to the timestamp or data acquisition sequence \cite{schonberger2014structure,aliakbarpour2015fast}. According to this principle, Cheng et al. \cite{cheng2022near} proposed a strategy to connect sequential images for image localization and stereo-pair dense matching, which uses the optical images sequentially acquired by UAV to achieve the real-time 3D reconstruction of disaster areas. For the latter, image match pairs are usually obtained according to camera mounting angles or onboard POS (position and orientation system) data. Using the projection center of images, Rupnik et al. \cite{rupnik2013automatic} searched the neighboring images close to the target image within the specified distance threshold. After acquiring the orientation data provided by the POS data of onboard navigation systems, image footprints on a specified elevation plane can be calculated, and image match pairs can be obtained through the pairwise intersection test between the image footprints \cite{jiang2017efficient,xu2016skeletal}. In the work of \cite{jiang2017efficient}, ground coverages of images are calculated by using POS data, and image match pairs are determined by judging the intersection of ground coverages. Although these methods have high efficiency, their accuracy depends on the used prior knowledge.

\subsection{Visual similarity-based methods}
\label{sec:2.2}

Compared with prior knowledge-based methods, these methods make match pairs selection using the images' content instead of prior knowledge. These methods can be grouped into two categories: the first is based on the number of matched correspondences, while the second uses the similarity score computed from image descriptors. For the former, two images are labeled as a valid match pair when the number of matches surpasses a threshold, such as the multi-scale strategy \cite{verykokou2018photogrammetry} and the preemptive matching strategy \cite{wu2013towards}. For the latter, images are quantified as descriptors, and the similarity score between two images is calculated as the distance between two descriptors. One of the most classic methods is vocabulary tree-based image retrieval \cite{qilarge,duan2015distributed}. Using a trained vocabulary tree, this method quantizes extracted local features into word frequency vectors, i.e., BoW (Bags-of-Words) vectors. The distance between the vectors represents the similarity score between the images \cite{jiang2019uav}. These methods can quickly obtain correct match pairs on small datasets, which is inefficient for large-scale datasets. In addition to the above-mentioned methods, neural network-based methods have been proposed recently. Yan et al. \cite{yan2021image} proposed a match pair selection method based on the GCN (Graph Convolutional Network) and used it to judge whether overlapping areas exist between images. This method performed remarkably well on challenging datasets from ambiguous and duplicated scenes. However, the efficiency is very low for high-resolution UAV images.

\subsection{Nearest neighbor searching}
\label{sec:2.3}

NN searching aims to find the vectors closest to the query vector from a large set of database vectors. In the context of match pair selection, the NN searching in vocabulary tree-based image retrieval is solved as an ANN searching problem, which determines the efficiency of image retrieval. In the literature, existing ANN searching methods can be divided into three categories, i.e., tree-based methods, hashing-based methods, and graph-based methods. Tree-based methods use a tree structure to partition the searching space, and KD-Tree is one of the most well-known data structures \cite{bentley1975multidimensional}, which has been used extensively for image retrieval algorithms \cite{huang2010video,hu2019high} and software packages, e.g., the COLMAP \cite{schonberger2016structure} and AliceVision \cite{griwodz2021alicevision}, because of the relative low dimension of used feature descriptors, such as the 128-dimension SIFT (Scale Invariant Feature Transform) descriptor. However, the efficiency of tree-based methods decreases dramatically for high-dimension vectors, which is not better than brute-force searching. To increase ANN searching efficiency, hashing-based methods convert continuous real-value vectors to discrete binary codes using hashing functions. In this category, LSH (Locality-Sensitive Hashing) attempts to hash similar vectors into the same cell with high probabilities \cite{indyk1998approximate}. Consequently, ANN searching can be executed in the cell that the query vector also falls in. Compared with the tree-based method, the hash operation reduces high-dimensional input vectors to low-dimensional terms by using a set of hash functions whose number is much smaller than the dimension of input vectors. This is useful to avoid the curse of dimensionality in tree-based methods. Due to their high efficiency, LSH-based methods have been used for large-scale image retrievals, such as web community and remote sensing images \cite{li2021recent}. These methods, however, have lower precision caused by the usage of binary hashing encoding as well as high memory consumption to store hashing functions. In contrast to splitting the searching space, graph-based methods create a graph data structure to organize database vectors and achieve efficient ANN searching based on graph indexing. NSW (Navigable Small World) \cite{malkov2014approximate} and HNSW (Hierarchical NSW) \cite{malkov2018efficient} are two typical graph-based methods. NSW adopts an approximation of the Delaunay graph, which has the same operation for vertex insertion and query. NSW can achieve efficient and accurate searching based on long-distance edges that are created at the beginning, which forms a small navigable world and reduces the number of hops. HNSW is an improved version of NSW, which builds a multi-layer structure to speed up ANN searching. In the work of \cite{liu2022efficient}, HNSW has been used to replace the KD-Tree in image retrieval, and good acceleration has been achieved in match pair selection. However, unacceptable time consumption is still required for processing large-scale UAV images due to a large number of local features.

\section{Methodology}
\label{sec:3}

\begin{figure}[ht!]
    \centering
    \includegraphics[width=0.5\textwidth]{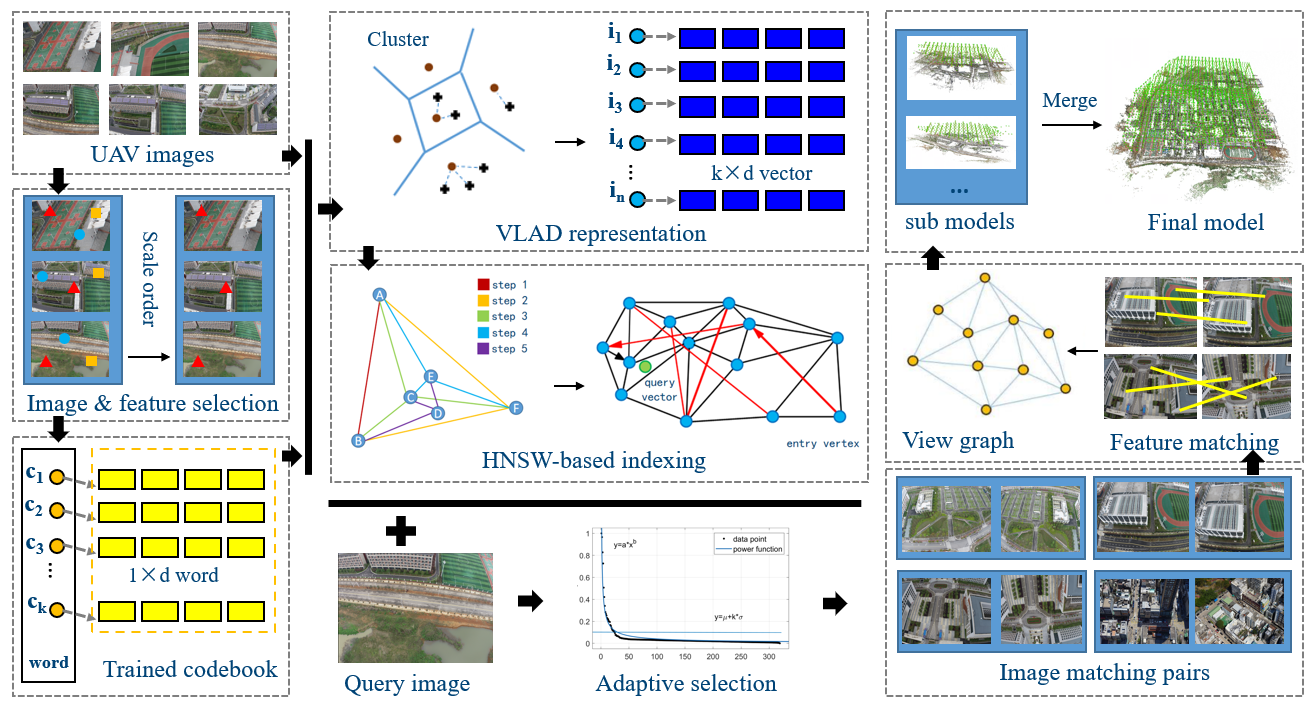}
    \caption{The proposed match pair retrieval workflow and parallel SfM solution.}
    \label{fig:fig1}
\end{figure}

This study proposes an efficient and accurate match pair retrieval method for large-scale UAV images and implements a parallel SfM solution guided by the view graph constructed from retrieved match pairs. The core idea is to use global descriptors for image representation and explores a graph indexing structure for the ANN searching of high-dimension vectors. The workflow of the complete SfM reconstruction is shown in Figure \ref{fig:fig1}, in which the inputs are UAV images without other auxiliary data. First, a codebook is trained online by selecting a subset of UAV images and scale-restricted features. Second, with the aid of the codebook, each image's local features are aggregated into a single high-dimension vector according to VLAD. Third, VLAD vectors are then indexed into an HNSW-based graph structure to achieve highly efficient ANN searching, and match pairs are retrieved based on the HNSW index and refined by using an adaptive selection strategy. Finally, after feature matching guided by the retrieved match pairs, a weighted view graph is constructed, which is used for the scene partition and parallel SfM reconstruction of large-scale UAV images.

\subsection{Vocabulary tree-based image retrieval}
\label{sec:3.1}

Vocabulary tree-base image retrieval mimics the text retrieval that encodes a document as a feature vector by using trained words and casts document searching as the distance calculation between feature vectors \cite{sivic2006video}. The most important techniques are the inverted file for the word-image indexing and the TF-IDF (Term Frequency and Inverse Document Frequency) for the weighting of similarity scores \cite{nister2006scalable}.

The workflow of vocabulary tree-based image retrieval consists of four major steps. First, local features with descriptors, e.g., SIFT, are extracted from training images; second, a vocabulary tree is hierarchically built from the extracted descriptors by using a clustering algorithm, e.g., the K-means, whose leaf nodes indicate the generated visual words; third, all images are indexed by searching the nearest visual word for all extracted feature descriptors, and an inverted file is simultaneously built for each visual word, which builds the indexing relationship between visual words and image features; finally, the same indexing operation is executed for an input query image, and the similarity score between the query and database images can be calculated by using their corresponding BoW vectors. Suppose there is a vocabulary with $V$ words, and each image can be represented by a BoW vector $v_d=(t_1,...,t_i,...,t_V)$. The component $t_i$ is calculated according to Equation \ref{eq:1}

\begin{equation}
	t_{i}=\frac{n_{id}}{n_{d}}\log\frac{N}{N_{i}}
	\label{eq:1}
\end{equation}
where $n_{id}$ and $n_d$ indicate the occurrence number of the word $i$ in the image $d$ and the total number of words in image $d$, respectively; $N_i$ is the number of images that contain word $i$, and $N$ is the total number of images in the database. The component $t_i$ includes two parts, i.e., the term frequency (TF) $n_{id}/n_d$ and the inverse document frequency (IDF) $\log(N/N_i)$, which indicate the occurrence frequency of the word $i$ in the image $d$ and the importance of word $i$ among database images. After generating the BoW vectors, the similarity score of any two images can be quantified as the dot production of corresponding BoW vectors.

With the increasing of involved database images, vocabulary tree-based image retrieval efficiency decreases dramatically. The main reason is building the inverted index. On the one hand, the high resolution of UAV images leads to a large number of extracted features that cause high computational costs in the ANN searching to build the inverted file; on the other hand, with the increasing of database images, a larger codebook with more visual words must be used to increase the discriminative power of BoW vectors, which further increases the burden in the ANN searching and subsequent similarity calculation. Therefore, considering these issues, this study proposes an efficient image retrieval solution that combines the VLAD descriptor and the HNSW indexing. The former aggregates local feature descriptors into a high-dimensional global vector using a very small codebook, which avoids the high computational costs in image indexing; the latter is utilized to accelerate the ANN searching for high-dimensional VLAD vectors. This study would integrate the proposed solution with a parallel SfM workflow for large-scale image orientation. The details are described in the following sections.

\subsection{Codebook generation considering image and feature redundancy}
\label{sec:3.2}

Local features are first detected from UAV images as training data. In recent years, UAV images have been capable of recording building facades and observing ground targets from multi-view directions. Due to the large differences in viewing directions and the obvious changes in illuminations and scales, feature matching becomes non-trivial for oblique UAV images \cite{jiang2018efficient}. Considering the issues of oblique UAV images, the SIFT algorithm extracts local features. In this study, to balance the accuracy and efficiency of subsequent match pair selection, 8,129 local features with the highest scales are extracted for each image, and the feature descriptors are represented as a vector with dimension = 128.

By using extracted local features, a codebook can be generated for the aggregation of local features to the VLAD descriptor. In general, there are two ways to generate a codebook, i.e., one for online generation for each dataset and the other for offline generation for all datasets. While the second way accelerates online processing without training an individual codebook, it cannot represent the characteristics of specified datasets and provide inferior performance on image retrieval. Therefore, the optimal way is to generate a codebook for an individual UAV dataset \cite{arandjelovic2013all}. However, it would be very time-consuming to generate a codebook because the large data volume and high spatial resolution of UAV images cause many descriptors. For UAV images, there are two kinds of redundancy. The first is the image redundancy due to the high overlap degree to ensure the success of subsequent image orientation; the second is the feature redundancy because of the high spatial resolution of UAV images. These two kinds of redundancy could be exploited to reduce the descriptor number in codebook training. On the one hand, the number of visual words for VLAD aggregation is extremely less than that for BoW indexing \cite{jegou2011aggregating}. A very coarse quantization of the descriptor space is required for VLAD aggregation. On the other hand, the characteristics of one image can be represented by a subset of features with large scales. Thus, this study proposes a random sampling strategy to select a subset $p$ of training images and a scale restriction strategy to select a subset $h$ of descriptors with large scales. Based on the work \cite{jiang2020efficient}, the parameter $p$ and $h$ are set as 20\% and 1500.

After selecting the training descriptors, the codebook with $k$ clusters is generated by using the K-means clustering algorithm \cite{araihierarchical}: 1) pick $k$ cluster centers randomly; 2) assign each descriptor to its nearest cluster center; 3) calculate the mean vector of each cluster and use it as the new cluster centers; 4) repeat steps 2) to 3) after a certain number of iteration times or reach the convergence condition of the algorithm. Based on the clustering algorithm, the $k$ cluster centers indicate the codebook $C=\{c_1,c_2,c_3,...,c_k\}$. The number of cluster centers $k$ is closely related to the performance of the match pair retrieval algorithm. On the one hand, the accuracy of match pair retrieval will be reduced when $k$ is too small; on the other hand, the generation of the codebook will consume more memory, and the efficiency of subsequent feature aggregation and image retrieval will be reduced when $k$ is too large. Thus, a proper $k$ is significant for match pair retrieval.

\subsection{Adaptive match pair retrieval via global descriptor and graph indexing}
\label{sec:3.3}

\subsubsection{Global descriptor from the aggregation of local features}
\label{sec:3.3.1}

Some solutions are designed for aggregating local features to global vectors, e.g., the BoW that counts the term frequency of words. However, the number of words in the trained codebook should increase simultaneously with the number of involved images. It would cause high time costs for large-scale image indexing. Instead of the term frequency counting, VLAD accumulates residuals between local feature descriptors and their corresponding cluster centers and achieves high discriminative power using a very small-size codebook. Based on the observation, this study uses VLAD to aggregate local features into global descriptors \cite{jegou2011aggregating}.

For $N$ extracted local features of an image, the VLAD descriptor is obtained by iterating feature descriptors assigned to the same cluster center and calculating the sum of the residuals between these feature descriptors and the cluster center. The final VLAD descriptor is a concatenation of residual vectors generated from all cluster centers. Supposing that there are $k$ cluster centers in the trained codebook $C$, the VLAD descriptor $v$ consists of $k$ vectors with the same dimension $d=128$ as the used SIFT descriptor. Therefore, the calculation of an element $v_{k,j}$ in the VLAD descriptor $v$ is presented by Equation \ref{eq:2}

\begin{equation}
	v_{k,j}=\sum_{i=1}^Na_k(d_i)(d_i(j)-c_k(j))
	\label{eq:2}
\end{equation}
where $j$ is the dimension index of feature descriptors, i.e., $j=1,2,...,d$; $a_k(d_i)$ is an indicator function: when the feature descriptor $d_i$ belongs to the visual word $c_k$, $a_k(d_i)=1$; otherwise, $a_k(d_i)=0$. Based on the formulation, an image is represented as a $k\times d$ VLAD descriptor. Compared with the BoW vector, the VLAD descriptor uses the residual vector to encode the input image. In order to generate the same dimension feature vector, extremely fewer visual words are required in the trained codebook, i.e., the ratio is the same as the dimension $d=128$ of the used descriptors. Besides, component-wise and global L2-normalization is sequentially conducted for the generated VLAD descriptors. Noticeably, the VLAD aggregation can be executed parallelly because it is independent for each clustering center.


\subsubsection{Match pair retrieval based on Graph-indexed global descriptors}
\label{sec:3.3.2}

Match pairs can be selected by the nearest neighbor searching between VLAD descriptors. Recently, graph-based solutions have attracted enough attention because of their high precision and promising efficiency when dealing with high-dimension descriptors. HNSW \cite{malkov2018efficient} is one of the well-known graph-based search algorithms, which is implemented based on the NSW (Navigable Small World) search method \cite{malkov2014approximate}. HNSW uses a hierarchical structure to build a vector index graph to increase retrieval efficiency, miming a coarse-to-fine searching strategy. The bottom layer includes all vertices, and the number of vertices decreases gradually from the bottom to up layers. In the retrieval stage, after the entry of the query vector, the HNSW index is used to search from top to bottom, which restricts the searching of the next nearest neighbor to the child nodes in the next layer. The nearest neighbors in the bottom layers are the retrieval results. Thus, HNSW is used in this study for high-dimension multi-VLAD vector indexing and match pair retrieval. The VLAD descriptors are first constructed into a graph structure $G=\{V, E\}$, in which $V$ and $E$ respectively represent the vertex set composed of VLAD descriptors and the edge set composed of their connection relationships. To achieve efficient indexing and retrieval, the maximum number of connections for each vertex is restricted to $M$, termed the friend number. This parameter $M$ influences the efficiency and precision of image retrieval.

\begin{figure}[ht!]
    \centering
    \includegraphics[width=0.5\textwidth]{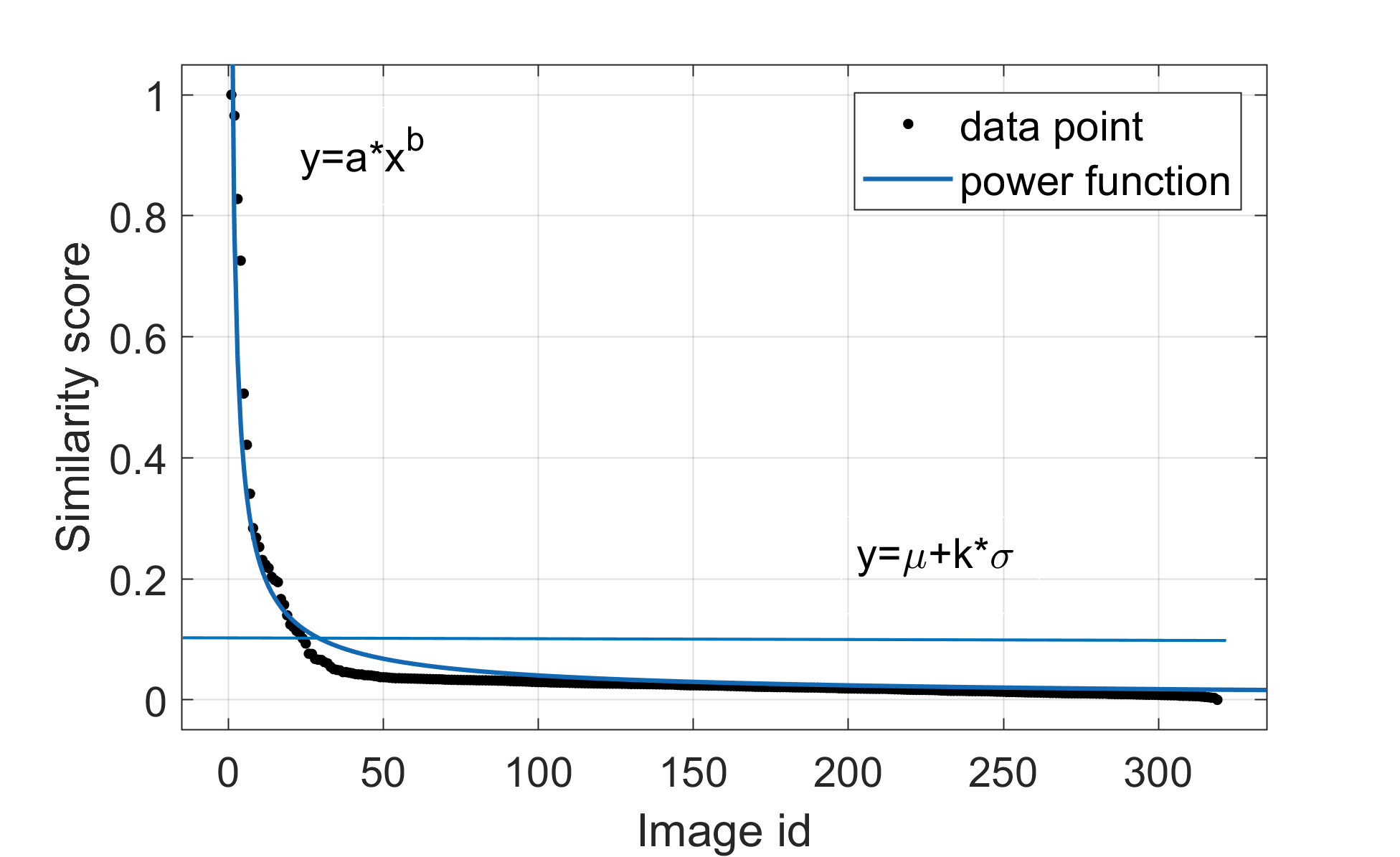}
    \caption{The distribution of similarity scores for one query image.}
    \label{fig:fig3}
\end{figure}

In match pair retrieval, the number of returned items should be specified well. The optimal value should adapt to the data acquisition configuration, mainly affected by the image overlap degree. It varies for each data acquisition and each UAV image. However, it is usually set as a fixed number or ratio in the classical image retrieval pipeline. In this study, an adaptive selection strategy has been adopted to select the number of retrieved images \cite{jiang2020efficient}. The core idea origins from the fact that images with larger overlap areas have higher similarity scores, and the similarity scores decrease dramatically with the decrease of overlap areas. However, image pairs without overlap areas have very small similarity scores, and at the same time, no obvious changes are observed from similarity scores, as illustrated in Figure \ref{fig:fig3}. Thus, the distribution of similarity scores is fitted well by using a power function with coefficients $a$ and $b$, as presented by Equation \ref{eq:3}

\begin{equation}
	y=a^*x^b
	\label{eq:3}
\end{equation}
where $x$ and $y$ indicate the image ids and similarity scores, respectively. Using the mean $\mu$ and standard deviation $\delta$ of similarity scores between one query and database images, a horizontal separation line $y=\mu+k\delta$ can be defined, and database images with similarity scores above the separation line are labeled as the retrieval results. Noticeably, in the HNSW-based image retrieval, the Euclidean distance instead of the similarity score has been returned. In this study, inverse linear normalization is used to calculate similarity scores. Suppose that $m$ items are retrieved with distance $D=\{d_1,d_2,d_3,...,d_m\}$, the similarity score is calculated based on Equation \ref{eq:4}

\begin{equation}
	s_i=\frac{d_{max}-d_i}{d_{max}-d_{min}}
	\label{eq:4}
\end{equation}
where $d_{min}$ and $d_{max}$ indicate the minimal and maximal values in $D$, respectively. Thus, this equation converts the Euclidean distance to the similarity score that ranges from 0 to 1. Besides, the separation line $y=\mu+k\delta$ is mainly influenced by the mean $\mu$ and standard deviation $\delta$. With the increase of used samples to fit the power function, the separation line $y$ would go down and retain more retrieved results. Thus, according to practical experiences, the number of used samples is set as 300 in this study.

\subsection{View graph construction from retrieved match pairs}
\label{sec:3.4}

False match pairs inevitably exist because of repetitive image patterns and non-optimal parameters in image retrieval. In this study, local feature matching and geometric verification are conducted to filter false matches. Guided by initial match pairs, local feature matching is performed by finding the nearest neighbors from two sets of features based on the Euclidean distance between feature descriptors, in which the cross-checking and ratio test have also been utilized. To further refine the initial matches, the epipolar geometry based on the Fundamental matrix is utilized to remove false matches, which can be robustly estimated in the RANSAC (Random Sampling Consensus) framework \cite{fischler1981paradigm}. Finally, the match pairs with the number of refined matches greater than 15 are retained.

A view graph can be created using the retained match pairs and their feature matches. In this study, the view graph is represented as an undirected weighted graph $G=\{V, E\}$, in which $V$ and $E$ indicate the vertex set and edge set, respectively \cite{jiang2022parallel}. Suppose that $I=\{i_i\}$ and $P=\{p_{ij}\}$ are respectively $n$ images and $m$ match pairs. The graph $G$ is constructed as follows: a vertex $v_i$ is added for each image $i_i$ and all vertices form the vertex set $V=\{v_i\}$; adding an edge $e_{ij}$ connecting vertex $v_i$ and vertex $v_j$ for each matched pair $p_{ij}$ and all edges form the edge set $E=\{e_{ij}\}$. To quantify the importance of match pairs, an edge weight $w_{ij}$ is assigned to the edge $e_{ij}$. In the context of SfM-based image orientation, the number of feature matches and their distribution over image planes directedly influence the overall performance. Thus, $w_{ij}$ is calculated by Equation \ref{eq:5}

\begin{equation}
	w_{ij}=R_{ew}\times w_{inlier}+(1-R_{ew})\times w_{overlap}
	\label{eq:5}
\end{equation}
where $R_{ew}$ is the weight ratio between $w_{inlier}$ and $w_{overlap}$, which is set as 0.5 similar to the work in \cite{jiang2022parallel}. $w_{inlier}$ is the weight item related to the number of feature matches; $w_{overlap}$ is the weight item related to the distribution of feature matches. These two items are calculated respectively according to Equations \ref{eq:6} and \ref{eq:7}

\begin{equation}
	w_{inlier} = \frac{\log(N_{inlier})}{\log(N_{max\_inlier)}}
	\label{eq:6}
\end{equation}

\begin{equation}
	w_{overlap}=\frac{CH_i+CH_j}{A_i+A_j}
	\label{eq:7}
\end{equation}
where $N_{inlier}$ and $N_{max\_inlier}$ indicate the number of matched correspondences of the match pair and the maximum number of matched correspondences among all match pairs; $CH_i$ and $CH_j$ represent the convex hull areas of feature matches over two images; $A_i$ and $A_j$ represent the areas of two image planes. In our study, the Graham-Andrew algorithm \cite{andrew1979another} is used to detect convex hulls of feature matches.

\subsection{Parallel SfM reconstruction guided by view graph}
\label{sec:3.5}

In this study, an incremental SfM is used to estimate camera poses and scene structures. Incremental SfM, however, suffers the problem of low efficiency due to the sequential registering of images and iterative local and global bundle adjustment. For large-scale scenes, this issue becomes very obvious and limits the applications of SfM in recent photogrammetric systems. To overcome the problem, this study adopts the divide-and-conquer strategy to split the large-size reconstruction into small-size sub-reconstructions. Thus, sub-reconstructions can be well addressed, and parallel techniques can also be utilized to improve efficiency. Figure \ref{fig:fig4} illustrates the basic principle of the designed parallel SfM solution \cite{jiang2022parallel}, which includes four major steps described as follows:

\begin{figure}[ht!]
    \centering
    \includegraphics[width=0.5\textwidth]{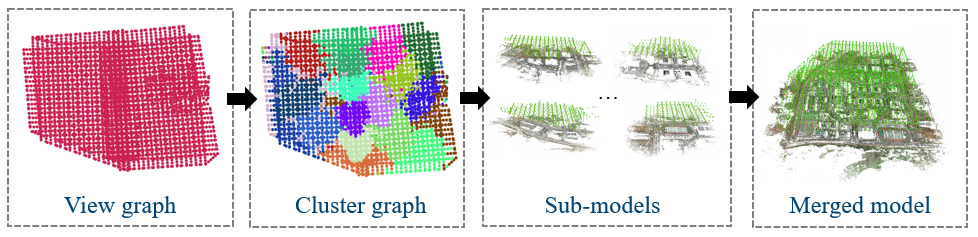}
    \caption{The workflow of parallel SfM reconstruction.}
    \label{fig:fig4}
\end{figure}

\begin{itemize}

\item 
First, after creating the view graph $G$, the scene is divided into small-size clusters $\{G_i\}$ with strong inner connections. The scene clustering is implemented through the NC (Normalized Cut) algorithm \cite{shi2000normalized}, which removes the edges with smaller weights and ensures the good connection of vertices in each cluster.

\item 
Second, an incremental SfM engine is then executed parallelly for each cluster $G_i$, which generates an individual model for each cluster. In this study, the well-known incremental SfM engine, COLMAP \cite{schonberger2016structure}, has been utilized to implement the parallel reconstruction of each cluster.

\item 
Third, cluster merging is performed by iteratively merging two sub-models, which convert individual models to an entire model in the same global coordinate system. In this step, the merging order is critical as it affects the robustness and precision of cluster merging. In this study, the number of common 3D points between models is used to sort the merging order, which has been calculated efficiently through a corresponding graph established between two clusters \cite{jiang2022parallel}.

\item 
Finally, a final global bundle adjustment is executed for the merged global model. Since the number of optimization parameters would be very large, a tie-point selection strategy is adopted to decrease the number of 3D points in BA optimization. As documented in \cite{chen2020graph}, tie-point selection is achieved based on four metrics, i.e., re-projection error, overlap degree, image coverage, and number limitation.
\end{itemize}

\subsection{Algorithm implementation}
\label{sec:3.6}

This study implements the solution of match pair retrieval and parallel SfM reconstruction using the C++ programming language, as presented in Algorithm \ref{alg:ParallelSfM}. In detail, for feature extraction, the SIFTGPU \cite{2013SiftGPU} library is used with default parameter setting; for the generation of the codebook, the Lloyd’s K-means cluster algorithm \cite{lloyd1982least} has been used; in addition, we have implemented an algorithm for the aggregation of SIFT features into VLAD descriptors and adopted the HNSW algorithm in the FAISS package \cite{johnson2019billion} for graph indexing; based on our previous work \cite{jiang2022parallel}, we have embedded the match pair retrieval and view graph construction method into the parallel SfM workflow, in which the software package ColMap \cite{schonberger2016structure} has been selected as the incremental SfM engine.


\begin{algorithm}[h!]
	\caption{Effcient Match Pair Retrieval and Parallel SfM}
	\label{alg:ParallelSfM}
	\hspace*{\algorithmicindent} \textbf{Input:} $n$ input images $I=\{i_i\}$ \\
	\hspace*{\algorithmicindent} \textbf{Output:} reconstructed model $M$
	\begin{algorithmic}[1]
		\Procedure{}{}
		\State Extract SIFT features $F=\{f_i\}$ for all input images $I$
            \State Train a codebook $VT$ by image and feature selection
            \State Generate VLAD descriptors $D=\{d_{vlad}\}$ based on $VT$
            \State Index $D$ into a graph $IDX$ based on the HNSW
            \State Retrieve match pairs $P=\{P_{ij}\}$ using index $IDX$
            \State Create view graph $G$ from refined match pairs $P_{refine}$
            \State Divide the view graph $G$ into sub-clusters $\{G_i\}$
            \State Execute parallel SfM to obtain sub-models $\{M_i\}$
            \State Merge sub-models $\{M_i\}$ into the global model $M$
		\EndProcedure
	\end{algorithmic}
	
\end{algorithm}

\section{Experiments and results}
\label{sec:4}

In the experiment, three UAV datasets have been collected to evaluate the performance of the proposed solution. First, according to the efficiency and precision of match pair selection, we analyze the influence of key parameters, i.e., the number of cluster centers $k$ for the codebook generation and the maximum number of neighboring vertices $M$ in HNSW. Second, we conduct the match pair selection and SfM-based 3D reconstruction of the three UAV datasets using the selected parameter setting. Third, we compared the proposed SfM solution with four well-known software packages, i.e., two open-source software packages ColMap \cite{schonberger2016structure} and DboW2 \cite{galvez2012bags} and two commercial software packages Agisoft Metashape and Pix4Dmapper, to evaluate the performance of match pair selection and SfM reconstruction. In the study, all experiments are executed on a Windows desktop computer with 64 GB memory, four Intel 2.40 GHz Xeon E5-2680 CPUs, and one 10 GB NVIDIA GeForce RTX 3080 graphics card.

\subsection{Test sites and datasets}
\label{sec:4.1}

\begin{figure}[ht!]
    \centering
    \includegraphics[width=0.5\textwidth]{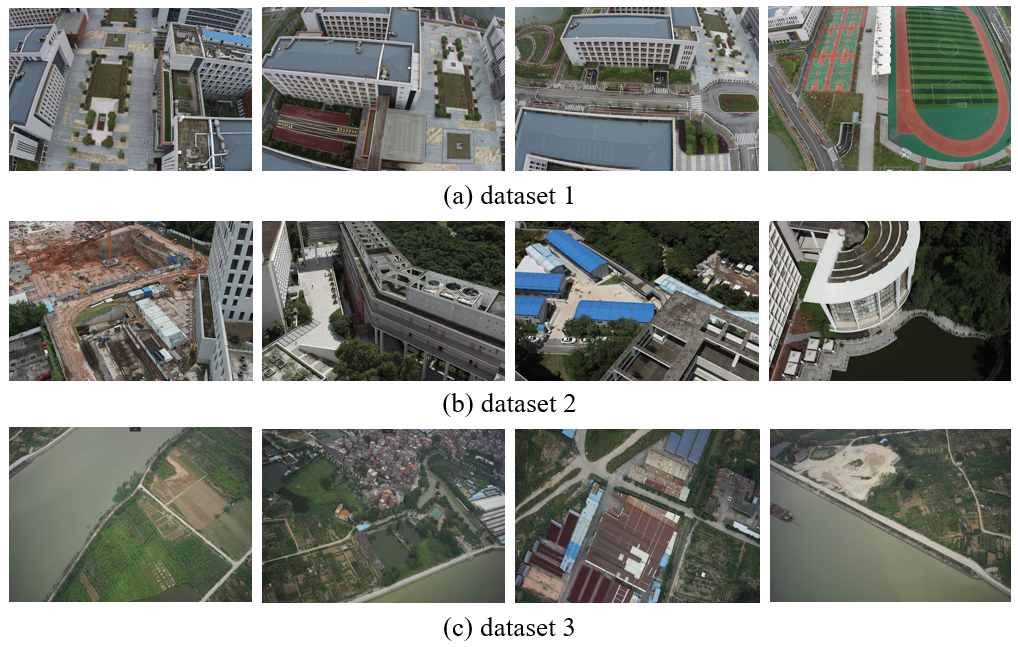}
    \caption{Sample images in the three test UAV datasets.}
    \label{fig:fig5}
\end{figure}

Three UAV datasets with different sizes are used for the performance evaluation. Figure \ref{fig:fig5} shows the sample images in each dataset, and the detailed information is listed in Table \ref{tab:table1}. The description of each dataset is presented as follows:

\begin{itemize}
\item 
The first dataset consists of 3,743 images taken from a university campus covered by dense and low-rise buildings. The dataset is captured by a DJI Phantom 4 RTK UAV equipped with one DJI FC6310R camera. The images with 5,472 by 3,648 pixels are collected under the flight height of 80 m, and the GSD (Ground Sample Distance) is approximately 2.6 cm.

\item
The second dataset includes 4,030 images taken from a complex university building. It is captured using a DJI M300 RTK UAV equipped with one DJI Zenmuse P1 camera with a dimension of 8,192 by 5,460 pixels. It is worth mentioning that this dataset has been collected based on the optimized views photogrammetric \cite{li2023optimized}, which adjusts camera viewpoints and directions according to the geometry of ground objects. The GSD is approximately 1.2 cm. For absolute orientation, 26 GCPs (Ground Control Points) were collected using a total station, whose nominal accuracy is about 0.8 and 1.5 cm in the horizontal and vertical directions.

\item
The third dataset is recorded by a penta-view oblique photogrammetric instrument equipped with five SONY ILCE 7R cameras with 6,000 by 4,000 pixels. Low-rise buildings and dense vegetation mainly cover this test site. In addition, a rive comes across the test site. Under the flight height of 87.1 m, a total number of 21,654 images has been collected with a GSD of 1.21 cm.
\end{itemize}

\begin{table}[ht!]
	\centering
	\caption{Detailed information of the three UAV datasets.}
	\label{tab:table1}
	\makebox[0.5\linewidth]{
		\begin{tabular}{l l l l}
			\toprule
			\textbf{Item Name} & \textbf{Dataset 1} & \textbf{Dataset 2} & \textbf{Dataset 3} \\
			\midrule
			UAV type & multi-rotor &  multi-rotor & multi-rotor \\
			Flight height (m) & 80 & - & 87.1 \\
			Camera mode & DJI FC6310R & DJI ZenmuseP1 & SONY ILCE 7R \\
			Camera number & 1 & 1 & 5 \\
			Focal length (mm) & 24 & 35 & 35 \\
			Camera angle ($^\circ$) & 0 & - & \begin{tabular}{@{}l@{}} nadir: 0 \\ oblique: 45/$-$45 \end{tabular} \\
			Number of images & 3,743 & 4,030 & 21,654 \\
			Image size (pixel)  & 5472 $\times$ 3648 & 8192 $\times$ 5460 & 6000 $\times$ 4000 \\
			GSD (cm) & 2.6 & 1.2 & 1.21 \\
			\bottomrule
		\end{tabular}
	}
\end{table}

\subsection{The influence of parameters K and M}
\label{sec:4.2}

For the proposed match pair retrieval solution, two critical parameters directly influence the efficiency and precision of image indexing and retrieval, i.e., the visual word number $k$ in the generation of the trained codebook and the friend number $M$ in the graph-based indexing. The former determines the dimension of the VLAD vectors; the latter determines the maximum number of connections of each vertex to others in the HNSW graph. Thus, this section analyzes their influence on retrieval efficiency and precision.

For the evaluation, dataset 1 has been selected, and two metrics are used for performance evaluation: retrieval efficiency and precision. The retrieval efficiency is the total time costs consumed in match pair selection; the retrieval precision is calculated as the ratio between the number of correct match pairs and the number of all match pairs. In this test, the retrieval time includes time costs in VLAD-based feature aggregation, HNSW-based graph construction, and image retrieval. To avoid the influence of the adaptive selection, the retrieval number is fixed as 30, and match pairs with at least 15 true matches are defined as positive results.

\begin{figure}[ht!]
	\centering
	\subfloat[]{\includegraphics[width=0.25\textwidth]{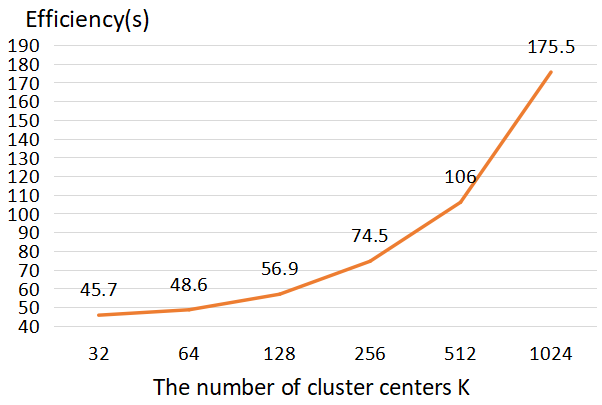}
		\label{fig:figure6a}}
	\subfloat[]{\includegraphics[width=0.25\textwidth]{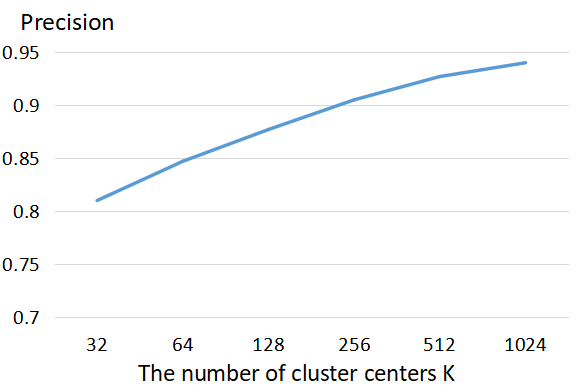}
		\label{fig:figure6b}}
	\caption{The influence of parameters $k$: (a) efficiency; and (b) precision.}
	\label{fig:fig6}
\end{figure}

For the analysis of the parameter $k$, the values of 32, 64, 128, 256, 512, and 1024 are tested. Figure \ref{fig:fig6} presents the statistical results of efficiency and precision in the match pair selection, in which Figure \ref{fig:figure6a} and Figure \ref{fig:figure6b} respectively indicate the efficiency and precision. It is clearly shown that with the increase of $k$, the time costs increase exponentially, from 45.7 seconds to 175.5 seconds, with the value ranging from 32 to 1024, respectively. The main reason is that a larger $k$ leads to more time costs in the nearest cluster center searching for VLAD feature aggregation and increases the dimension of generated VLAD descriptors, which further poses a burden in HNSW graph indexing and retrieval. On the contrary, we can observe that the retrieval precision increases linearly with the increase of the parameter $k$, which increases from 0.81 to 0.94 within the specified span. To balance efficiency and precision, the parameter $k$ is set as 256 in the following tests.

\begin{figure}[ht!]
	\centering
	\subfloat[]{\includegraphics[width=0.25\textwidth]{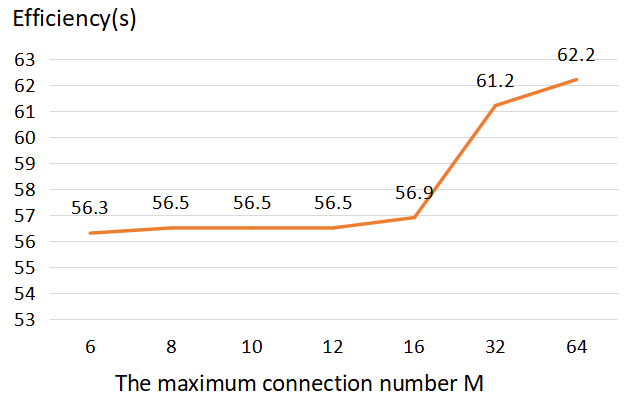}
		\label{fig:figure7a}}
	\subfloat[]{\includegraphics[width=0.25\textwidth]{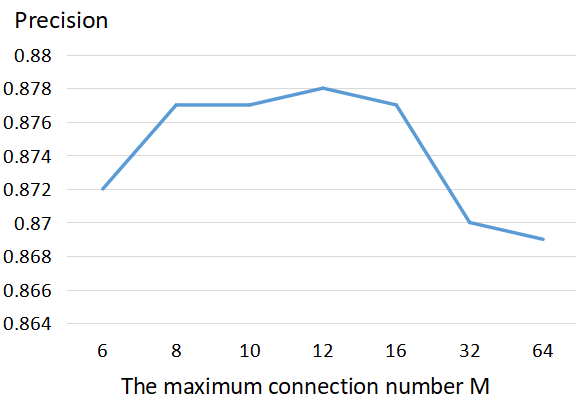}
		\label{fig:figure7b}}
	\caption{The influence of parameters $M$: (a) efficiency; and (b) precision.}
	\label{fig:fig7}
\end{figure}

For the analysis of the parameter $M$, the values of 6, 8, 10, 12, 16, 32, and 64 are used, and the statistical results are presented in Figure \ref{fig:fig7}. We can see that: (1) the changing trend of retrieval efficiency in Figure \ref{fig:figure7a} can be divided into two parts. In the first part, the retrieval efficiency is almost constant with the value $M$ increasing from 6 to 16; in the second part, the retrieval efficiency decreases dramatically with the value $M$ increasing from 16 to 64; (2) the changing trend of retrieval precision in Figure \ref{fig:figure7b} can be separated into three stages. In the first stage, the retrieval precision increases obviously with the value $M$ increasing from 6 to 8; in the second stage, the retrieval precision keeps constant within the value range from 8 to 16; in the third stage, the retrieval precision decreases gradually within the value range from 16 to 64. It is worth mentioning that $k$ has a greater impact on retrieval efficiency than $M$ because most time costs are spent in VLAD aggregation. Besides, $M$ affects the number of valid NN neighbors that can be retrieved. Considering that at least 300 valid NN neighbors should be retrieved in the adaptive selection, the parameter $M$ is set as 32 in the following tests.

\subsection{Match pairs selection and 3D reconstruction}
\label{sec:4.3}

\subsubsection{Match pairs selection by the proposed retrieval method}
\label{sec:4.3.1}

\begin{figure}[ht!]
	\centering
	\subfloat[]{\includegraphics[width=0.28\textwidth]{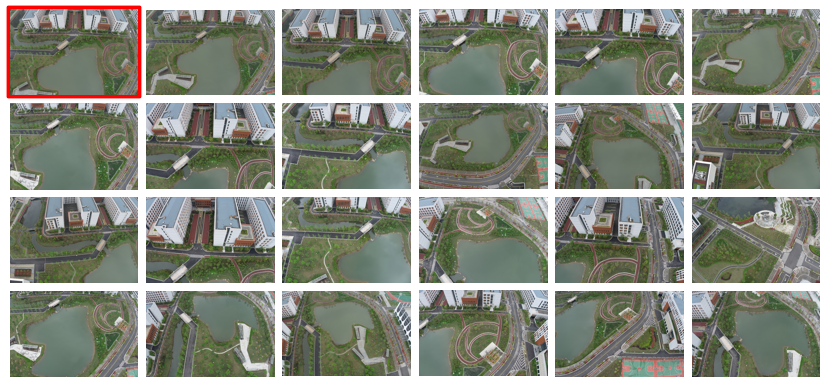}
		\label{fig:figure8a}}
	\subfloat[]{\includegraphics[width=0.22\textwidth]{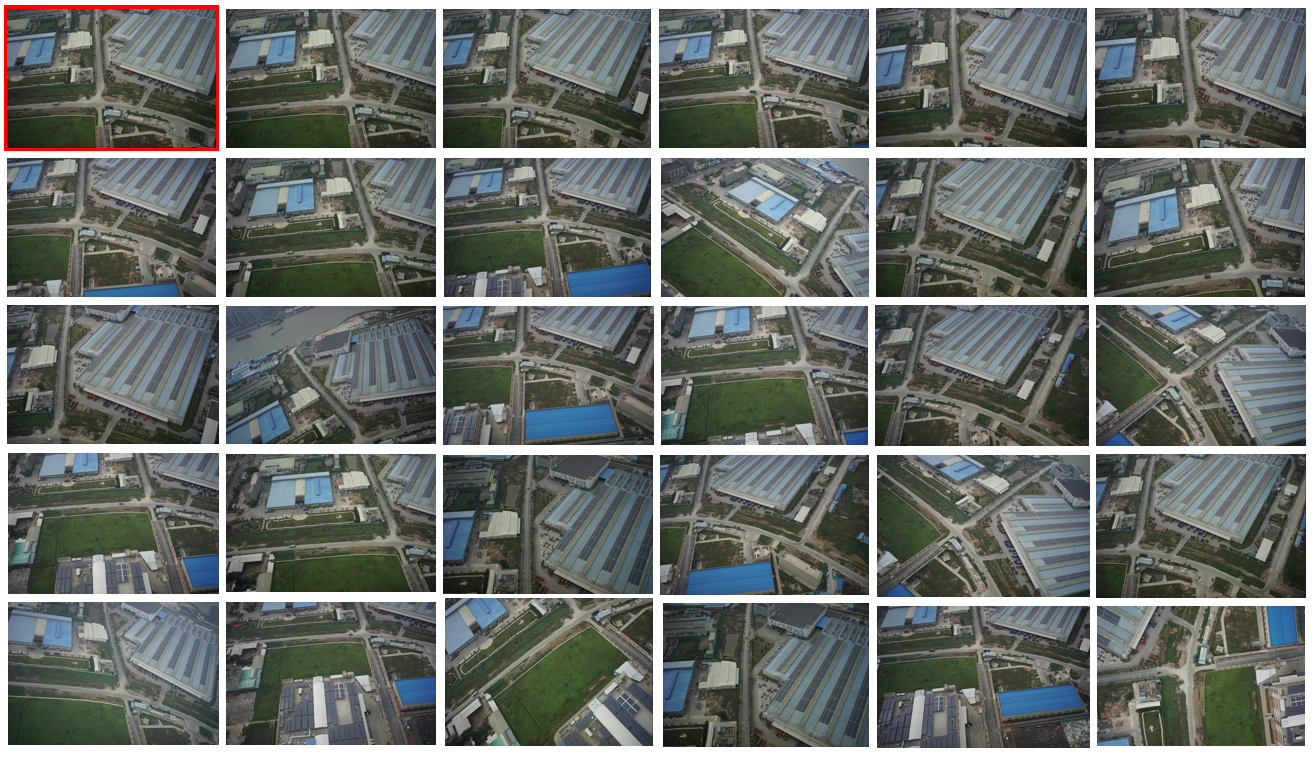}
		\label{fig:figure8b}}
	\caption{The illustration of match pair selection for (a) dataset 1; and (b) dataset 3. The images with and without red boundary boxes are the query and retrieved images, respectively.}
	\label{fig:fig8}
\end{figure}

By using the selected parameters $k$ and $M$, the performance of match pair selection is first evaluated. Similarly, retrieval efficiency and precision are used as the metrics for performance evaluation. Table \ref{tab:table2} lists the statistical results of match pair selection. It is clearly shown that high retrieval precision has been achieved for the three datasets, which are 90.1\%, 89.9\%, and 94.4\% for the three datasets, respectively. It ensures that a very large proportion of selected match pairs are overlapped images. Figure \ref{fig:fig8} shows the results of our method to retrieve similar images for two sample images from datasets 1 and 3. It can be seen that all the retrieved images are true positive results. In addition, the time costs of match pair selection are 2.5 mins, 2.6 mins, and 12.4 mins for the three datasets, respectively, which achieves the average time costs of approximately 0.040 secs, 0.039 secs, and 0.034 secs for match pair selection. Thus, we can conclude that the proposed solution can achieve linear time complexity in image indexing and retrieval and process large-scale UAV datasets for efficient match pair selection.

\begin{table}[ht!]
	\centering
	\caption{The statistical results of match pair selection for the three datasets. The values in the bracket indicate the time cost of codebook generation.}
	\label{tab:table2}
	\makebox[0.5\linewidth]{
		\begin{tabular}{l l l l}
			\toprule
			\textbf{Metric} & \textbf{Dataset 1} & \textbf{Dataset 2} & \textbf{Dataset 3} \\
			\midrule
			Efficiency (min) & 2.5(0.9) &  2.6(1.0) & 12.4(2.8) \\
			Precision (\%) & 90.1 & 89.9 & 94.4 \\
			
			\bottomrule
		\end{tabular}
	}
\end{table}

\subsubsection{Parallel 3D reconstruction guided by the weighted view graph}
\label{sec:4.3.2}

\begin{figure*}[ht!]
	\centering
	\subfloat[]{\includegraphics[width=0.3\textwidth]{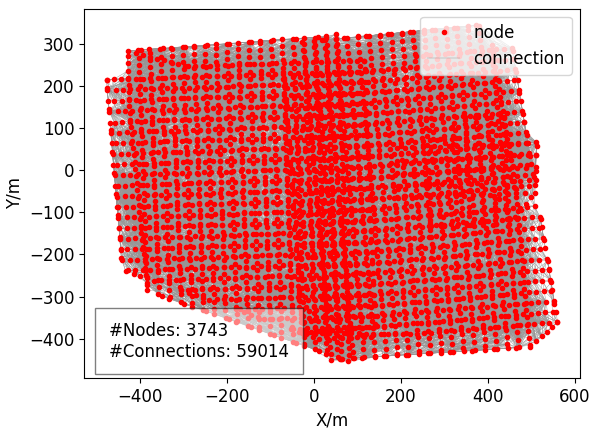}
		\label{fig:figure9a}}
	\subfloat[]{\includegraphics[width=0.3\textwidth]{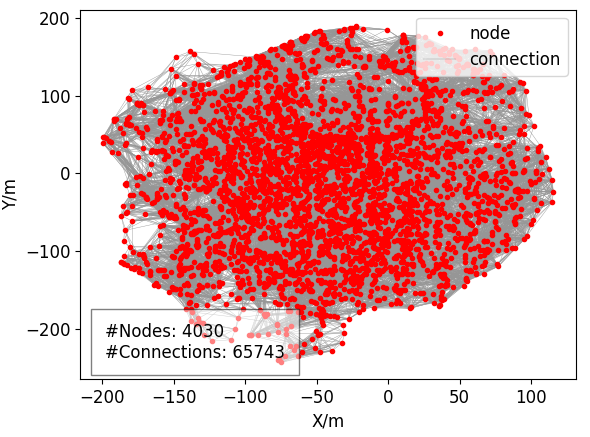}
		\label{fig:figure9b}}
        \subfloat[]{\includegraphics[width=0.3\textwidth]{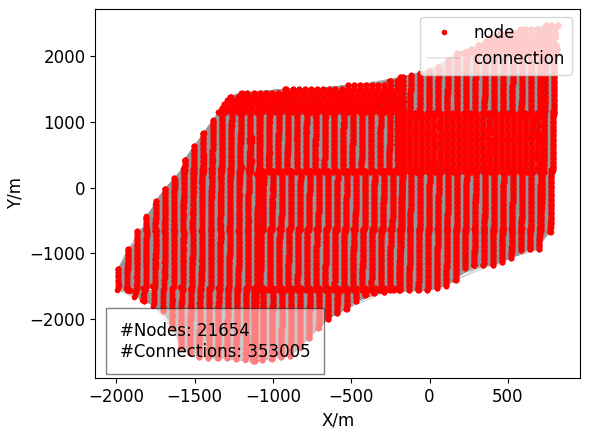}
		\label{fig:figure9c}}
	\caption{The topological connection network: (a) view graph of dataset 1; (b) view graph of dataset 2; (c) view graph of dataset 3. The red circles and gray lines indicate image positions and match pairs, respectively.}
	\label{fig:fig9}
\end{figure*}

The selected match pairs are then used to guide feature matching. In this study, feature matching is achieved by searching approximate nearest neighbors, refined based on the widely used ratio test and cross-checking. The initial matches are then verified by the epipolar constraint implemented by the estimation of the fundamental matrix within the framework of RANSAC. In this study, the threshold of ratio-test is set as 0.8 as the default value in the SIFTGPU library, and the maximum distance threshold is configured as 1.0 pixels to ensure the high inlier ratio of feature matching. Using feature matching results, a view graph represented as an undirected weighted graph can be constructed for each dataset, whose vertices and edges represent images and their connection relationships, respectively. As presented in Figure \ref{fig:fig9}, three view graphs are created for the three UAV datasets, in which vertices and edges are rendered by red dots and gray lines, respectively. It is shown that there are 59,014, 65,743, and 353,005 match pairs selected from the three datasets, respectively. The dense edges between vertices indicate a strong connection between images, which ensures the success of SfM-based image orientation.

\begin{figure*}[ht!]
	\centering
	\subfloat[]{\includegraphics[width=0.3\textwidth]{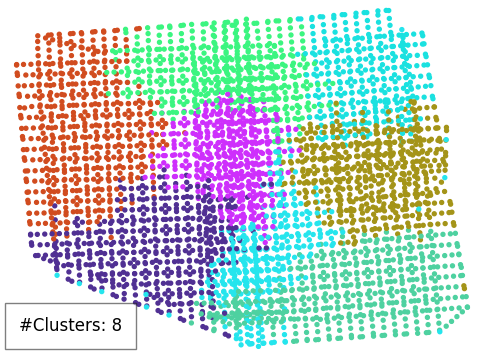}
		\label{fig:figure10a}} 
	\subfloat[]{\includegraphics[width=0.3\textwidth]{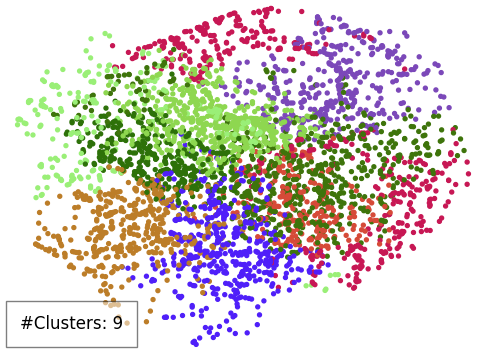}
		\label{fig:figure10c}}
    \subfloat[]{\includegraphics[width=0.3\textwidth]{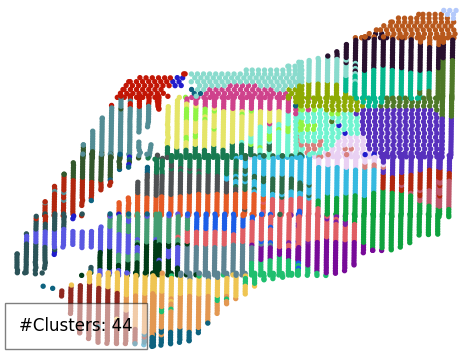}
		\label{fig:figure10e}} \\
    \subfloat[]{\includegraphics[width=0.3\textwidth]{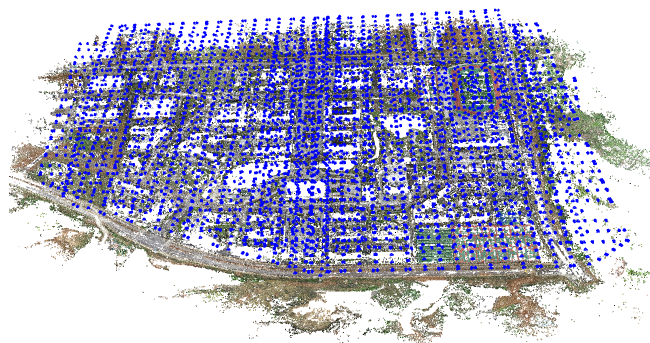}
		\label{fig:figure10b}}
    \subfloat[]{\includegraphics[width=0.3\textwidth]{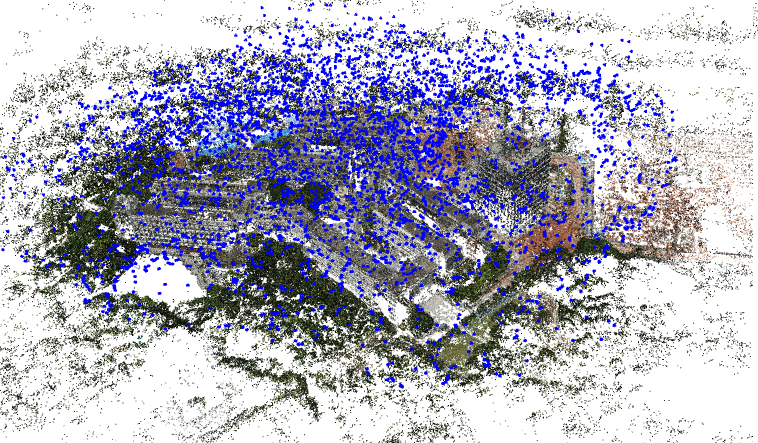}
		\label{fig:figure10d}}
    \subfloat[]{\includegraphics[width=0.3\textwidth]{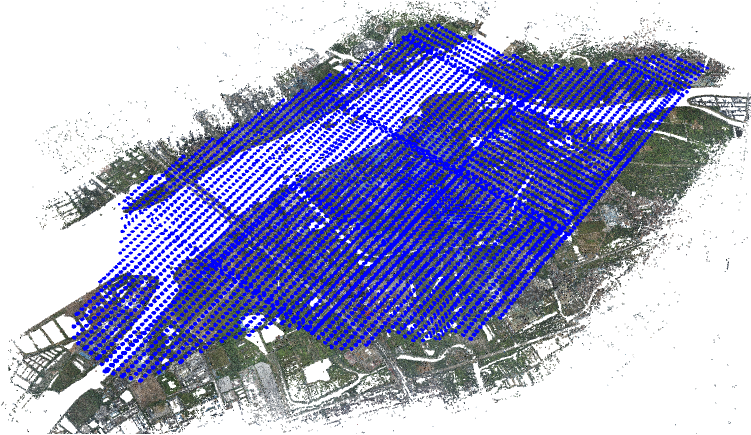}
		\label{fig:figure10f}}
	\caption{The illustration of scene clustering and parallel reconstruction for the three datasets: (a), (b), and (c) are scene clustering with the cluster size 500 for datasets 1, 2, and 3, respectively; (d), (e), and (f) are reconstructed 3D points for the three datasets, respectively.}
	\label{fig:fig10}
\end{figure*}

To achieve the parallel SfM reconstruction, the entire view graph is then divided into small sub-clusters with strong inner-edge connections. In the proposed parallel SfM workflow, the normalized cut algorithm is utilized for scene clustering, and the largest size of each sub-cluster is set as 500. The scene partition results are illustrated in Figure \ref{fig:figure10a}, Figure \ref{fig:figure10c}, and Figure \ref{fig:figure10e}. We can see 8, 9, and 44 sub-clusters generated for the three datasets. Each cluster is represented by an identical color, which verifies the compact connections within each cluster. Based on the sub-clusters, parallel SfM is executed to create the sub-reconstructions that are finally merged into the entire reconstruction. Table \ref{tab:table3} shows the statistical results of 3D reconstruction, in which the metrics precision and completeness refer to the re-projection error of BA optimization and the numbers of oriented images and reconstructed 3D points. We can see that the precision of the three datasets are 0.542 pixels, 0.668 pixels, and 0.752 pixels, respectively, and almost all images are oriented successfully, whose numbers are 3,724, 4,029, and 21,568, respectively. For the visualization, Figure \ref{fig:figure10b}, Figure \ref{fig:figure10d}, and Figure \ref{fig:figure10f} shows the reconstructed 3D points from the three datasets. It is shown that the reconstructed 3D points can cover the whole test site. Thus, the proposed solution can create stable view graphs to achieve parallel SfM.

\begin{table}[ht!]
	\centering
	\caption{The statistical results of match pair selection for the three datasets. The values in the bracket indicate the time cost of codebook generation.}
	\label{tab:table3}
	\makebox[0.5\linewidth]{
		\begin{tabular}{l l l l}
			\toprule
			\textbf{Metric} & \textbf{Dataset 1} & \textbf{Dataset 2} & \textbf{Dataset 3} \\
			\midrule
			Precision (pixel) & 0.542 &  0.668 & 0.752 \\
			Completeness & 928,745 (3,724) & 1,518,474 (4,029) & 8,921,339 (21,568) \\
			
			\bottomrule
		\end{tabular}
	}
\end{table}

\subsection{Performance comparison with the other software packages}
\label{sec:4.4}


\subsubsection{Match pair selection}
\label{sec:4.4.1}

\begin{figure}[ht!]
	\centering
	\subfloat[]{\includegraphics[height=0.15\textwidth]{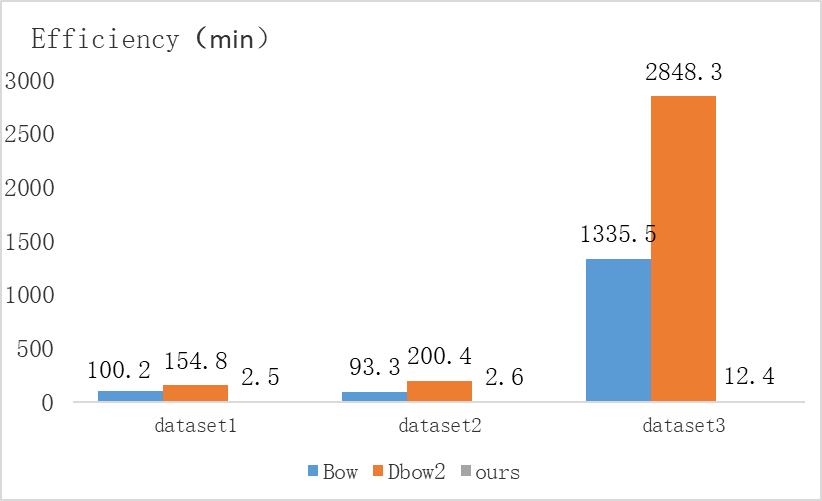}
		\label{fig:figure11a}}
	\subfloat[]{\includegraphics[height=0.15\textwidth]{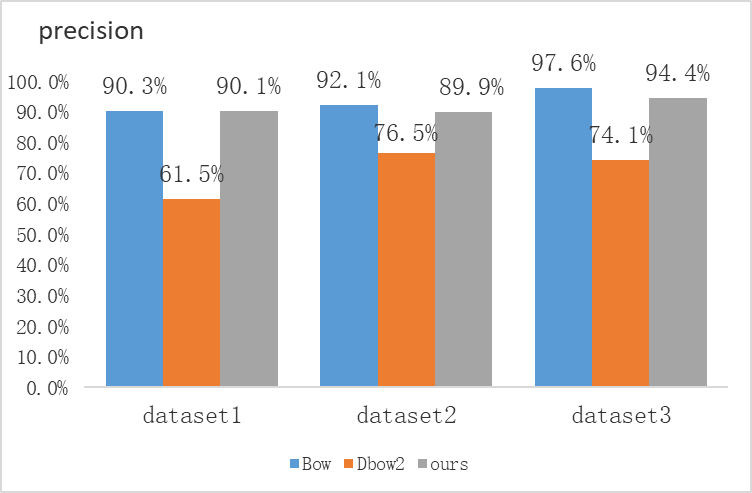}
		\label{fig:figure11b}}
	\caption{Comparison of match pair selection: (a) efficiency and (b) precision.}
	\label{fig:fig11}
\end{figure}

The proposed solution is compared with the BoW retrieval method in ColMap and the Dbow2 retrieval method to evaluate the performance in match pair selection. The statistic result is presented in Figure \ref{fig:fig11}. It is clearly shown that compared with BoW and Dbow2, the proposed solution achieves the highest efficiency, whose time costs are 2.5 min, 2.6 min, and 12.4 min for the three datasets. Especially for dataset 3, the time costs of Bow and Dbow2 reach 1335.5 mins and 2848.3 mins, respectively, which is unacceptable in practice. By observing the results presented in Figure \ref{fig:figure11b}, we can see that BoW almost achieves the highest precision, which is 90.3\%, 92.1\%, and 97.6\% for the three datasets, respectively. The proposed solution ranks second with a precision of 90.1\%, 89.9\%, and 94.4\% for the three datasets, which are higher than Dbow2. In conclusion, compared with BoW, the proposed solution can achieve comparable precision with the speedup ratios ranging from 36 to 108 for the three UAV datasets.

\subsubsection{SfM-based reconstruction}
\label{sec:4.4.2}

To evaluate the performance in the workflow of SfM-based reconstruction, the proposed solution is further compared with two commercial software packages Agisoft Metashape and Pix4Dmapper. Agisoft Metashape uses multi-scale matching and GNSS data for match pair selection; Pix4Dmapper provides a vocabulary tree-based image retrieval. In this test, camera intrinsic parameters are calibrated and fixed in SfM, and the match pairs selected from Bow and Dbow2 are fed into the proposed parallel SfM for reconstruction. Besides, 26 GCPs in the second dataset are used to evaluate geo-referencing accuracy. In the following tests, the metric efficiency indicates the time costs in SfM reconstruction without feature matching.

Table \ref{tab:table4} presents the statistical results of SfM reconstruction without GCPs. It is shown that BoW, Dbow2, and the proposed solution have almost the same efficiency because of using the same SfM engine. Although Metashape and Pix4Dmapper can achieve the reconstruction of datasets 1 and 2, their efficiency is lower, which further verifies the advantage of the parallel SfM workflow. Noticeably, Metashape and Pix4Dmapper fail to reconstruct dataset 3 since the large data volume causes the out-of-memory error in reconstruction. Considering the metric precision, it is shown that Pix4Dmapper achieves the highest performance, which BoW, Dbow2, and the proposed solution follow. For metric completeness, we can see that comparable performance can be observed from the evaluated software packages except for Pix4Dmapper. This is mainly caused by the relatively low precision of image retrieval.


\begin{table*}[t!]
	\centering
	\caption{The statistical results of SfM-based reconstruction without GCPs. The values in the bracket indicate the number of connected images.}
	\makebox[\linewidth]{
		\begin{tabular}{lrrrr}
			\toprule
			\textbf{Metric}               & \textbf{Method} & \textbf{Dataset 1} & \textbf{Dataset2} & \textbf{Dataset 3} \\
			\midrule
			\multirow{5}{*}{\begin{tabular}[c]{@{}l@{}}Efficiency\\ (mins)\end{tabular}}  & BoW & 32.9 & 145.2 & 1,753 \\
			& DBoW2          & 31.1               & 125.05            & 1,445             \\
			& Metashape     & 50.0              & 186.0            & -             \\
			& Pix4Dmapper            & 298.2               & 636.4            & -             \\
                & ours            & 32.5               & 144.0            & 1,778             \\
			\midrule
			\multirow{5}{*}{\begin{tabular}[c]{@{}l@{}}Precision\\ (pixels)\end{tabular}} & BoW & 0.542 & 0.667 & 0.766 \\
			& DBoW2          & 0.490               & 0.645            & 0.782             \\
			& Metashape     & 0.957              & 1.140            & -             \\
			& Pix4Dmapper            & 0.318               & 0.327            & -             \\
                & ours            & 0.542               & 0.668            & 0.752             \\
			\midrule
			\multirow{5}{*}{Completeness} & BoW & 1,001,797 (3,716) & 1,507,983 (4,029) & 9,253,968 (21,647) \\
			& DBoW2          & 925,530 (3,720)               & 1,506,702 (4,027)            & 9,047,089 (21,625)             \\
			& Metashape     & 1,764,717 (3,741)              & 1,536,021 (4,030)            & -             \\
			& Pix4Dmapper            & 468,254 (3,620)               & 726,366 (3,909)            & -             \\
                & ours            & 928,745 (3,724)               & 1,518,474 (4,029)            & 8,921,339 (21,568)             \\
			\bottomrule
		\end{tabular}
		\label{tab:table4}
	}
\end{table*}


Absolute bundle adjustment with GCPs is further executed to evaluate the geo-referencing accuracy of reconstructed models. In this test, three GCPs that are evenly distributed over test site 2 are utilized for the geo-referencing of SfM reconstructed models, and the others are used as check points (CPs). For the performance evaluation, two metrics, i.e., mean and std.dev. of CPs residuals are used in this test. In addition, Pix4dMapper has been selected as a baseline for commercial software packages.

\begin{figure*}[ht!]
	\centering
	\subfloat[]{\includegraphics[width=0.3\textwidth]{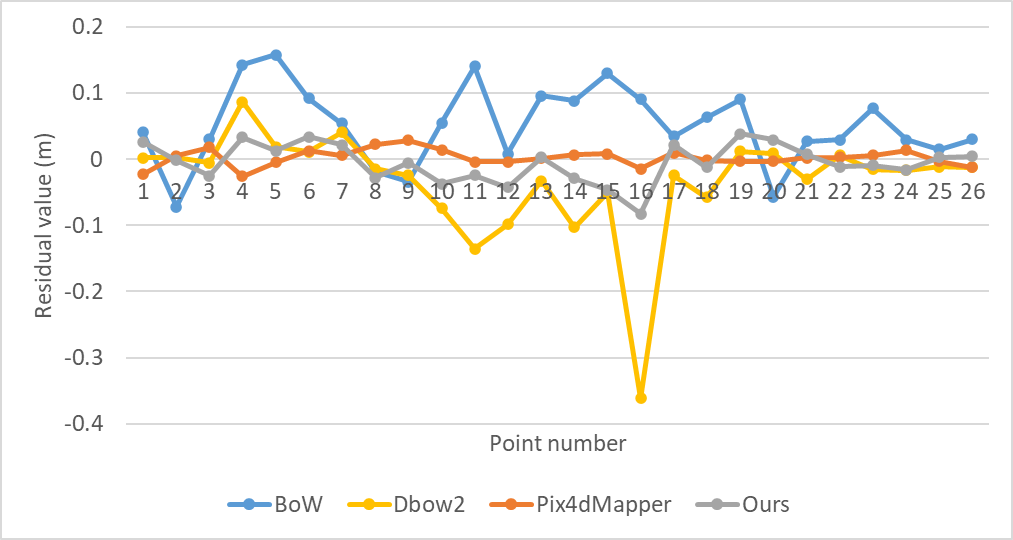}
		\label{fig:figure13a}}
	\subfloat[]{\includegraphics[width=0.3\textwidth]{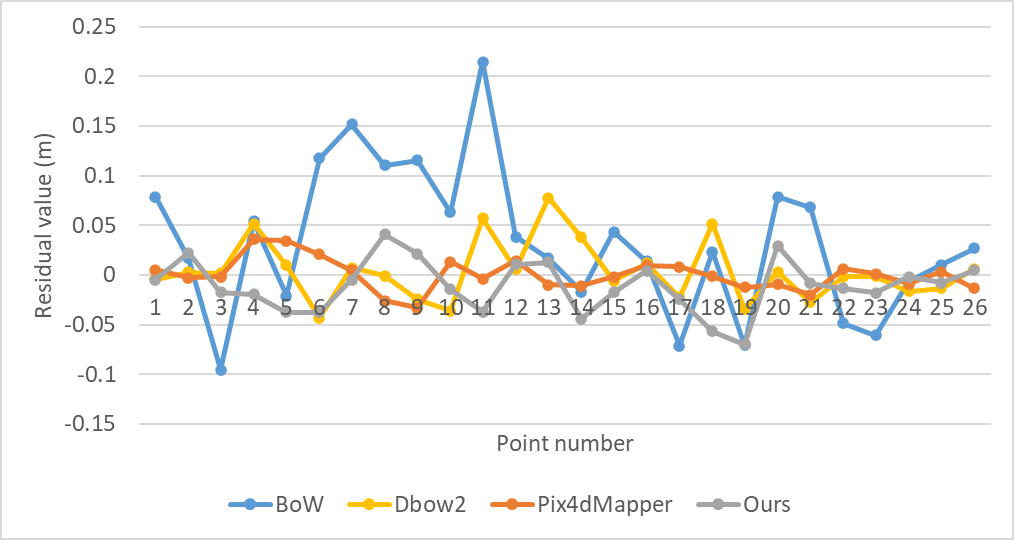}
		\label{fig:figure13b}}
    \subfloat[]{\includegraphics[width=0.3\textwidth]{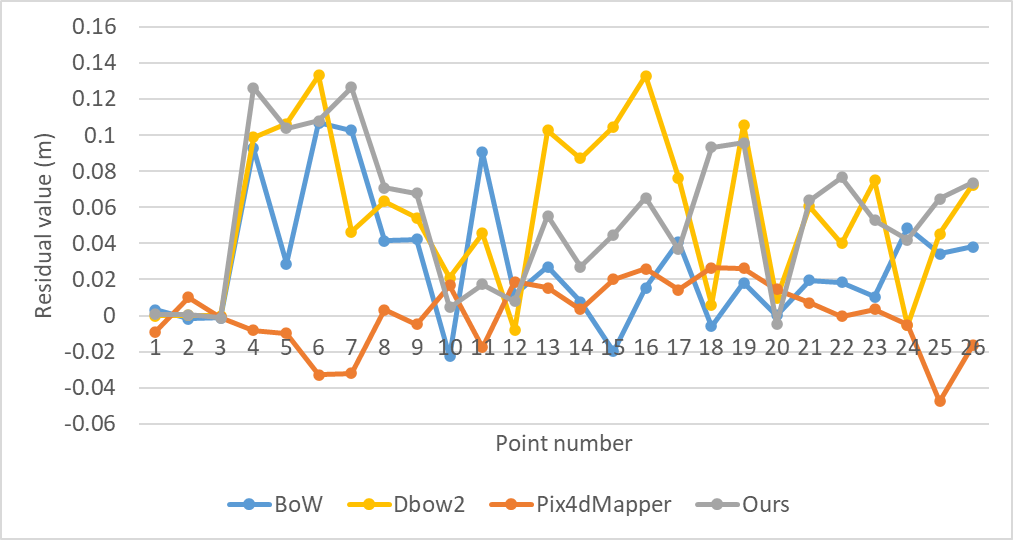}
		\label{fig:figure13c}}
	\caption{The residual plot in the X, Y, and Z directions for dataset 2: (a) the residual plot in the X direction; (b) the residual plot in the Y direction; (c) the residual plot in the Z direction.}
	\label{fig:fig13}
\end{figure*}

Table \ref{tab:table5} presents the statistical results of absolute BA. It is shown that among all evaluated software packages, Pix4dMapper achieves the highest accuracy with the std.dev. of 0.013 cm, 0.016 cm, and 0.019 cm in the X, Y, and Z directions, respectively. Although BoW ranks second in the vertical direction with the std.dev. of 0.036 cm, its horizontal accuracy is lower than the proposed solution with the std.dev. of 0.029 cm and 0.026 cm in the X and Y directions, respectively, which can also be verified by the residual plot presented in Figure \ref{fig:figure13a} and Figure \ref{fig:figure13b}. Due to the low precision of match pair selection, the geo-referencing accuracy of Dbow2 is the lowest in the X and Z directions, as shown in Figure \ref{fig:figure13a} and Figure \ref{fig:figure13c}. Thus, we can conclude that the proposed solution can provide necessary and accurate match pairs to achieve reliable SfM reconstruction with obviously high efficiency.

\begin{table}[ht!]
	\centering
	\caption{The statistical results of absolute BA with GCPs for dataset 3.}
	\begin{tabular}{lllllll}
		\toprule
		\multirow{2}{*}{Method} & \multicolumn{3}{l}{Mean (m)} & \multicolumn{3}{l}{Std.dev. (m)} \\
		\cline{2-7}
		& $|X|$ & $|Y|$ & $|Z|$ & $|X|$ & $|Y|$ & $|Z|$ \\
		\midrule
		BoW & 0.066 & 0.063 & 0.033 & 0.059 & 0.074 & 0.036  \\
		DBoW2 & 0.048 & 0.021 & 0.058 & 0.081 & 0.030 & 0.044  \\
		Pix4dMapper & 0.010 & 0.012 & 0.015 & 0.013 & 0.016 & 0.019  \\
		Ours & 0.023 & 0.022 & 0.055 & 0.029 & 0.026 & 0.040 \\
		\bottomrule
	\end{tabular}
	\label{tab:table5}
\end{table}

\section{Conclusions}
\label{sec:5}

In this paper, we proposed a workflow that integrates match pair retrieval and parallel SfM reconstruction to achieve the efficient and accurate 3D reconstruction of large-scale UAV images. The core idea of match pair selection is to aggregate many local features into high-dimensional global vectors that can then be indexed through a graph-based structure for efficient ANN searching. Guided by the selected match pairs, a weighted view graph is created to achieve the parallel SfM through graph clustering and sub-model merging. The tests demonstrate that the proposed workflow can significantly accelerate match pair selection with a speedup ratio of tens and hundreds of times and increase the efficiency of SfM-based reconstruction with comparative results.

In this study, some observations and possible limitations have also been observed. First, the precision of match pair selection is dramatically influenced by the number of words in the codebook generated through K-means clustering, as shown in Section \ref{sec:4.2}. At the same time, a large K would also decrease the image retrieval efficiency. Thus, it is non-trivial to trade precision and efficiency, especially for large-scale datasets. Second, the hand-crafted local features, i.e., SIFT, are adopted for image retrieval of their high tolerance to scale and viewpoint changes. However, deep learning-based feature detectors have attracted enough attention in the fields of image retrieval \cite{arandjelovic2016netvlad} and feature matching \cite{jiang2021unmanned} due to the excellent ability of representation learning. Therefore, it is rational to use learned descriptors to enhance the image retrieval and feature matching algorithm in the proposed workflow. Third, only the CPU is used in the implemented algorithm, which can be further accelerated using the GPU parallel computing technique. In future research, we will conduct more tests on selecting high-quality match pairs with high efficiency by exploiting learned feature descriptors and the GPU acceleration technique.

\section*{Acknowledgments}
This research was funded by the National Natural Science Foundation of China (Grant No. 42001413), the Open Research Fund from the Guangdong Laboratory of Artificial Intelligence and Digital Economy (SZ) (Grant No. GML-KF-22-08), the Open Research Project of The Hubei Key Laboratory of Intelligent Geo-Information Processing (Grant No. KLIGIP-2021B11), and the Provincial Natural Science Foundation of Hunan (Grant No. 2023JJ30232).

\bibliographystyle{IEEEtran}
\nocite{*}
\bibliography{bare_jrnl}

\begin{thebibliography}{10}
\providecommand{\url}[1]{#1}
\csname url@samestyle\endcsname
\providecommand{\newblock}{\relax}
\providecommand{\bibinfo}[2]{#2}
\providecommand{\BIBentrySTDinterwordspacing}{\spaceskip=0pt\relax}
\providecommand{\BIBentryALTinterwordstretchfactor}{4}
\providecommand{\BIBentryALTinterwordspacing}{\spaceskip=\fontdimen2\font plus
\BIBentryALTinterwordstretchfactor\fontdimen3\font minus
  \fontdimen4\font\relax}
\providecommand{\BIBforeignlanguage}[2]{{%
\expandafter\ifx\csname l@#1\endcsname\relax
\typeout{** WARNING: IEEEtran.bst: No hyphenation pattern has been}%
\typeout{** loaded for the language `#1'. Using the pattern for}%
\typeout{** the default language instead.}%
\else
\language=\csname l@#1\endcsname
\fi
#2}}
\providecommand{\BIBdecl}{\relax}
\BIBdecl

\bibitem{jiang2021unmanned}
S.~Jiang, W.~Jiang, and L.~Wang, ``Unmanned aerial vehicle-based
  photogrammetric 3d mapping: A survey of techniques, applications, and
  challenges,'' \emph{IEEE Geoscience and Remote Sensing Magazine}, vol.~10,
  no.~2, pp. 135--171, 2021.

\bibitem{li2023optimized}
Q.~Li, H.~Huang, W.~Yu, and S.~Jiang, ``Optimized views photogrammetry:
  Precision analysis and a large-scale case study in qingdao,'' \emph{IEEE
  Journal of Selected Topics in Applied Earth Observations and Remote Sensing},
  vol.~16, pp. 1144--1159, 2023.

\bibitem{jiang2019uav}
S.~Jiang and W.~Jiang, ``Uav-based oblique photogrammetry for 3d reconstruction
  of transmission line: Practices and applications.'' \emph{International
  Archives of the Photogrammetry, Remote Sensing \& Spatial Information
  Sciences}, 2019.

\bibitem{jiang2017uav}
S.~Jiang, W.~Jiang, W.~Huang, and L.~Yang, ``Uav-based oblique photogrammetry
  for outdoor data acquisition and offsite visual inspection of transmission
  line,'' \emph{Remote Sensing}, vol.~9, no.~3, p. 278, 2017.

\bibitem{colomina2014unmanned}
I.~Colomina and P.~Molina, ``Unmanned aerial systems for photogrammetry and
  remote sensing: A review,'' \emph{ISPRS Journal of photogrammetry and remote
  sensing}, vol.~92, pp. 79--97, 2014.

\bibitem{jiang2020efficient}
S.~Jiang and W.~Jiang, ``Efficient match pair selection for oblique uav images
  based on adaptive vocabulary tree,'' \emph{ISPRS Journal of Photogrammetry
  and Remote Sensing}, vol. 161, pp. 61--75, 2020.

\bibitem{schonberger2016structure}
J.~L. Schonberger and J.-M. Frahm, ``Structure-from-motion revisited,'' in
  \emph{Proceedings of the IEEE conference on computer vision and pattern
  recognition}, 2016, pp. 4104--4113.

\bibitem{nikolakopoulos2017uav}
K.~G. Nikolakopoulos, K.~Soura, I.~K. Koukouvelas, and N.~G. Argyropoulos,
  ``Uav vs classical aerial photogrammetry for archaeological studies,''
  \emph{Journal of Archaeological Science: Reports}, vol.~14, pp. 758--773,
  2017.

\bibitem{wischounig2015resource}
D.~Wischounig-Strucl and B.~Rinner, ``Resource aware and incremental mosaics of
  wide areas from small-scale uavs,'' \emph{Machine Vision and Applications},
  vol.~26, pp. 885--904, 2015.

\bibitem{chen2020graph}
Y.~Chen, S.~Shen, Y.~Chen, and G.~Wang, ``Graph-based parallel large scale
  structure from motion,'' \emph{Pattern Recognition}, vol. 107, p. 107537,
  2020.

\bibitem{cui2021view}
H.~Cui, T.~Shi, J.~Zhang, P.~Xu, Y.~Meng, and S.~Shen, ``View-graph
  construction framework for robust and efficient structure-from-motion,''
  \emph{Pattern Recognition}, vol. 114, p. 107712, 2021.

\bibitem{aliakbarpour2015fast}
H.~AliAkbarpour, K.~Palaniappan, and G.~Seetharaman, ``Fast structure from
  motion for sequential and wide area motion imagery,'' in \emph{Proceedings of
  the IEEE International Conference on Computer Vision Workshops}, 2015, pp.
  34--41.

\bibitem{schonberger2014structure}
J.~L. Sch{\"o}nberger, F.~Fraundorfer, and J.~M. Frahm, ``Structure-from-motion
  for mav image sequence analysis with photogrammetric applications,''
  \emph{The International Archives of Photogrammetry, Remote Sensing and
  Spatial Information Sciences}, vol.~40, no.~3, p. 305, 2014.

\bibitem{jiang2018efficient}
S.~Jiang and W.~Jiang, ``Efficient sfm for oblique uav images: From match pair
  selection to geometrical verification,'' \emph{Remote Sensing}, vol.~10,
  no.~8, p. 1246, 2018.

\bibitem{xu2016skeletal}
Z.~Xu, L.~Wu, M.~Gerke, R.~Wang, and H.~Yang, ``Skeletal camera network
  embedded structure-from-motion for 3d scene reconstruction from uav images,''
  \emph{ISPRS journal of photogrammetry and remote sensing}, vol. 121, pp.
  113--127, 2016.

\bibitem{lowe2004distinctive}
D.~G. Lowe, ``Distinctive image features from scale-invariant keypoints,''
  \emph{International journal of computer vision}, vol.~60, pp. 91--110, 2004.

\bibitem{hou2023learning}
Q.~Hou, R.~Xia, J.~Zhang, Y.~Feng, Z.~Zhan, and X.~Wang, ``Learning visual
  overlapping image pairs for sfm via cnn fine-tuning with photogrammetric
  geometry information,'' \emph{International Journal of Applied Earth
  Observation and Geoinformation}, vol. 116, p. 103162, 2023.

\bibitem{nister2006scalable}
D.~Nister and H.~Stewenius, ``Scalable recognition with a vocabulary tree,'' in
  \emph{2006 IEEE Computer Society Conference on Computer Vision and Pattern
  Recognition (CVPR'06)}, vol.~2.\hskip 1em plus 0.5em minus 0.4em\relax Ieee,
  2006, pp. 2161--2168.

\bibitem{jiang2022leveraging}
S.~Jiang, W.~Jiang, and B.~Guo, ``Leveraging vocabulary tree for simultaneous
  match pair selection and guided feature matching of uav images,'' \emph{ISPRS
  Journal of Photogrammetry and Remote Sensing}, vol. 187, pp. 273--293, 2022.

\bibitem{galvez2012bags}
D.~G{\'a}lvez-L{\'o}pez and J.~D. Tardos, ``Bags of binary words for fast place
  recognition in image sequences,'' \emph{IEEE Transactions on Robotics},
  vol.~28, no.~5, pp. 1188--1197, 2012.

\bibitem{cheng2022near}
M.-L. Cheng, M.~Matsuoka, W.~Liu, and F.~Yamazaki, ``Near-real-time gradually
  expanding 3d land surface reconstruction in disaster areas by sequential
  drone imagery,'' \emph{Automation in Construction}, vol. 135, p. 104105,
  2022.

\bibitem{rupnik2013automatic}
E.~Rupnik, F.~Nex, and F.~Remondino, ``Automatic orientation of large blocks of
  oblique images,'' \emph{The International Archives of Photogrammetry, Remote
  Sensing and Spatial Information Sciences}, vol.~40, no. Part 1, p.~W1, 2013.

\bibitem{jiang2017efficient}
S.~Jiang and W.~Jiang, ``Efficient structure from motion for oblique uav images
  based on maximal spanning tree expansion,'' \emph{ISPRS Journal of
  Photogrammetry and Remote Sensing}, vol. 132, pp. 140--161, 2017.

\bibitem{verykokou2018photogrammetry}
S.~Verykokou and C.~Ioannidis, ``A photogrammetry-based structure from motion
  algorithm using robust iterative bundle adjustment techniques.'' \emph{ISPRS
  Annals of Photogrammetry, Remote Sensing \& Spatial Information Sciences},
  vol.~4, 2018.

\bibitem{wu2013towards}
C.~Wu, ``Towards linear-time incremental structure from motion,'' in \emph{2013
  International Conference on 3D Vision-3DV 2013}.\hskip 1em plus 0.5em minus
  0.4em\relax IEEE, 2013, pp. 127--134.

\bibitem{qilarge}
J.~QI, J.~ZHAO, Y.~XIE, and X.-n. CHEN, ``Large-s cale image retrieval
  method.''

\bibitem{duan2015distributed}
H.~Duan, Y.~Peng, G.~Min, X.~Xiang, W.~Zhan, and H.~Zou, ``Distributed
  in-memory vocabulary tree for real-time retrieval of big data images,''
  \emph{Ad Hoc Networks}, vol.~35, pp. 137--148, 2015.

\bibitem{yan2021image}
S.~Yan, M.~Zhang, S.~Lai, Y.~Liu, and Y.~Peng, ``Image retrieval for
  structure-from-motion via graph convolutional network,'' \emph{Information
  Sciences}, vol. 573, pp. 20--36, 2021.

\bibitem{bentley1975multidimensional}
J.~L. Bentley, ``Multidimensional binary search trees used for associative
  searching,'' \emph{Communications of the ACM}, vol.~18, no.~9, pp. 509--517,
  1975.

\bibitem{huang2010video}
K.-Y. Huang, Y.-M. Tsai, C.-C. Tsai, and L.-G. Chen, ``Video stabilization for
  vehicular applications using surf-like descriptor and kd-tree,'' in
  \emph{2010 IEEE International Conference on Image Processing}.\hskip 1em plus
  0.5em minus 0.4em\relax IEEE, 2010, pp. 3517--3520.

\bibitem{hu2019high}
L.~Hu and S.~Nooshabadi, ``High-dimensional image descriptor matching using
  highly parallel kd-tree construction and approximate nearest neighbor
  search,'' \emph{Journal of Parallel and Distributed Computing}, vol. 132, pp.
  127--140, 2019.

\bibitem{griwodz2021alicevision}
C.~Griwodz, S.~Gasparini, L.~Calvet, P.~Gurdjos, F.~Castan, B.~Maujean,
  G.~De~Lillo, and Y.~Lanthony, ``Alicevision meshroom: An open-source 3d
  reconstruction pipeline,'' in \emph{Proceedings of the 12th ACM Multimedia
  Systems Conference}, 2021, pp. 241--247.

\bibitem{indyk1998approximate}
P.~Indyk and R.~Motwani, ``Approximate nearest neighbors: towards removing the
  curse of dimensionality,'' in \emph{Proceedings of the thirtieth annual ACM
  symposium on Theory of computing}, 1998, pp. 604--613.

\bibitem{li2021recent}
X.~Li, J.~Yang, and J.~Ma, ``Recent developments of content-based image
  retrieval (cbir),'' \emph{Neurocomputing}, vol. 452, pp. 675--689, 2021.

\bibitem{malkov2014approximate}
Y.~Malkov, A.~Ponomarenko, A.~Logvinov, and V.~Krylov, ``Approximate nearest
  neighbor algorithm based on navigable small world graphs,'' \emph{Information
  Systems}, vol.~45, pp. 61--68, 2014.

\bibitem{malkov2018efficient}
Y.~A. Malkov and D.~A. Yashunin, ``Efficient and robust approximate nearest
  neighbor search using hierarchical navigable small world graphs,'' \emph{IEEE
  transactions on pattern analysis and machine intelligence}, vol.~42, no.~4,
  pp. 824--836, 2018.

\bibitem{liu2022efficient}
S.~Liu, S.~Jiang, Y.~Liu, W.~Xue, and B.~Guo, ``Efficient sfm for large-scale
  uav images based on graph-indexed bow and parallel-constructed ba
  optimization,'' \emph{Remote Sensing}, vol.~14, no.~21, p. 5619, 2022.

\bibitem{sivic2006video}
J.~Sivic and A.~Zisserman, ``Video google: Efficient visual search of videos,''
  \emph{Toward category-level object recognition}, pp. 127--144, 2006.

\bibitem{arandjelovic2013all}
R.~Arandjelovic and A.~Zisserman, ``All about vlad,'' in \emph{Proceedings of
  the IEEE conference on Computer Vision and Pattern Recognition}, 2013, pp.
  1578--1585.

\bibitem{jegou2011aggregating}
H.~J{\'e}gou, F.~Perronnin, M.~Douze, J.~S{\'a}nchez, P.~P{\'e}rez, and
  C.~Schmid, ``Aggregating local image descriptors into compact codes,''
  \emph{IEEE transactions on pattern analysis and machine intelligence},
  vol.~34, no.~9, pp. 1704--1716, 2011.

\bibitem{araihierarchical}
K.~Arai and A.~Ridho, ``Hierarchical k-means: an algorithm for centroids
  initialization for k-means.''

\bibitem{fischler1981paradigm}
M.~A. Fischler and R.~C. Bolles, ``A paradigm for model fitting with
  applications to image analysis and automated cartography (reprinted in
  readings in computer vision, ed. ma fischler,'' \emph{Comm. ACM}, vol.~24,
  no.~6, pp. 381--395, 1981.

\bibitem{jiang2022parallel}
S.~Jiang, Q.~Li, W.~Jiang, and W.~Chen, ``Parallel structure from motion for
  uav images via weighted connected dominating set,'' \emph{IEEE Transactions
  on Geoscience and Remote Sensing}, vol.~60, pp. 1--13, 2022.

\bibitem{andrew1979another}
A.~M. Andrew, ``Another efficient algorithm for convex hulls in two
  dimensions,'' \emph{Information Processing Letters}, vol.~9, no.~5, pp.
  216--219, 1979.

\bibitem{shi2000normalized}
J.~Shi and J.~Malik, ``Normalized cuts and image segmentation,'' \emph{IEEE
  Transactions on pattern analysis and machine intelligence}, vol.~22, no.~8,
  pp. 888--905, 2000.

\bibitem{2013SiftGPU}
C.~Wu, ``Siftgpu: A gpu implementation of scale invariant feature transform
  sift,'' 2013.

\bibitem{lloyd1982least}
S.~Lloyd, ``Least squares quantization in pcm,'' \emph{IEEE transactions on
  information theory}, vol.~28, no.~2, pp. 129--137, 1982.

\bibitem{johnson2019billion}
J.~Johnson, M.~Douze, and H.~J{\'e}gou, ``Billion-scale similarity search with
  gpus,'' \emph{IEEE Transactions on Big Data}, vol.~7, no.~3, pp. 535--547,
  2019.

\bibitem{arandjelovic2016netvlad}
R.~Arandjelovic, P.~Gronat, A.~Torii, T.~Pajdla, and J.~Sivic, ``Netvlad: Cnn
  architecture for weakly supervised place recognition,'' in \emph{Proceedings
  of the IEEE conference on computer vision and pattern recognition}, 2016, pp.
  5297--5307.

\bibitem{cui2017hsfm}
H.~Cui, X.~Gao, S.~Shen, and Z.~Hu, ``Hsfm: Hybrid structure-from-motion,'' in
  \emph{Proceedings of the IEEE conference on computer vision and pattern
  recognition}, 2017, pp. 1212--1221.

\bibitem{shiwei2016online}
H.~Shiwei, L.~Jing, Y.~Tao, L.~Zhaoyang, Z.~Fangbing, and W.~Lisong, ``Online
  real-time image retrieval based on large-scale vocabulary tree,'' in
  \emph{2016 IEEE 13th International Conference on Signal Processing
  (ICSP)}.\hskip 1em plus 0.5em minus 0.4em\relax IEEE, 2016, pp. 953--956.

\bibitem{jegou2010aggregating}
H.~J{\'e}gou, M.~Douze, C.~Schmid, and P.~P{\'e}rez, ``Aggregating local
  descriptors into a compact image representation,'' in \emph{2010 IEEE
  computer society conference on computer vision and pattern
  recognition}.\hskip 1em plus 0.5em minus 0.4em\relax IEEE, 2010, pp.
  3304--3311.

\bibitem{zhou2020offsite}
X.~Zhou, K.~Xie, K.~Huang, Y.~Liu, Y.~Zhou, M.~Gong, and H.~Huang, ``Offsite
  aerial path planning for efficient urban scene reconstruction,'' \emph{ACM
  Transactions on Graphics (TOG)}, vol.~39, no.~6, pp. 1--16, 2020.

\bibitem{chu2021remote}
J.~Chu, L.~Li, and X.~Xiao, ``Remote sensing image retrieval by multi-scale
  attention-based cnn and product quantization,'' in \emph{2021 40th Chinese
  Control Conference (CCC)}.\hskip 1em plus 0.5em minus 0.4em\relax IEEE, 2021,
  pp. 8292--8297.

\bibitem{jegou2010product}
H.~Jegou, M.~Douze, and C.~Schmid, ``Product quantization for nearest neighbor
  search,'' \emph{IEEE transactions on pattern analysis and machine
  intelligence}, vol.~33, no.~1, pp. 117--128, 2010.

\bibitem{jiang2021learned}
S.~Jiang, W.~Jiang, B.~Guo, L.~Li, and L.~Wang, ``Learned local features for
  structure from motion of uav images: A comparative evaluation,'' \emph{IEEE
  Journal of Selected Topics in Applied Earth Observations and Remote Sensing},
  vol.~14, pp. 10\,583--10\,597, 2021.

\bibitem{muja2009fast}
M.~Muja and D.~G. Lowe, ``Fast approximate nearest neighbors with automatic
  algorithm configuration.'' \emph{VISAPP (1)}, vol.~2, no. 331-340, p.~2,
  2009.

\end{thebibliography}

\end{document}